\pdfoutput=1

\documentclass[10pt,letterpaper]{article}
  \usepackage[T1]{fontenc}
  \usepackage{newtxtext}  %
  \usepackage{helvet}
  \usepackage{courier}

\usepackage[hyphens]{url}  %
\usepackage{graphicx} %
\usepackage{natbib}  %
\usepackage{caption} %
\usepackage{booktabs}
\usepackage{amsmath,amssymb,amsthm}
\usepackage{stmaryrd}
\usepackage{tikz}
\usepackage{subcaption}

\usepackage{geometry}
\usepackage{xcolor}   %
\usepackage{placeins} %
\usepackage{microtype}      %
\usepackage[hidelinks]{hyperref}
\hypersetup{%
  pdftitle={Factor Rank: A Canonical Algebraic Generalization of Tree Decision Diagrams},%
  pdfauthor={Alexis de Colnet, Stefan Szeider},%
  pdfkeywords={knowledge compilation, tree decision diagrams, tensor networks, canonical form, factor rank}}
\usepackage{enumitem}
\AtBeginDocument{%
  \let\cite\citep%
  \bibliographystyle{abbrvnat}%
}

\newcommand{\wlandscape}{0.51}

\newcommand{\wfmtA}{0.48}
\newcommand{\wscaling}{0.60}

\newcommand{\suppl}{the Zenodo repository}
\newcommand{\supplb}{see the Zenodo repository}

\newtheorem{theorem}{Theorem}
\newtheorem{lemma}[theorem]{Lemma}
\newtheorem{proposition}[theorem]{Proposition}

\theoremstyle{definition}
\newtheorem{definition}[theorem]{Definition}
\newtheorem{example}[theorem]{Example}
\theoremstyle{remark}

\usepackage{thmtools}

\newcommand{\Q}{\mathbb{Q}}
\newcommand{\rk}{\operatorname{rank}}
\newcommand{\fw}{\operatorname{fw}}
\newcommand{\fr}{\operatorname{fr}}
\newcommand{\var}{\operatorname{var}}
\newcommand{\wt}[1]{\operatorname{wt}(#1)}
\newcommand{\HWB}{\mathrm{HWB}}
\newcommand{\MUX}{\mathrm{MUX}}
\definecolor{figmux}{HTML}{1B6CA8}%
\definecolor{fighwb}{HTML}{C8590A}%

\newcommand{\SUMBIT}{\mathrm{SUMBIT}}

\usepackage{soul}
\usepackage{nicematrix}

\usepackage{atddtikz}
\tikzset{
  atdd ttab/.append style={nodes={font=\small}},
  atdd dimlab/.append style={font=\small},
}

\title{Factor Rank: A Canonical Algebraic Generalization\\ of Tree Decision Diagrams}

\author{%
  \begin{tabular}{c@{\hspace{2.5em}}c}
    Alexis de Colnet & Stefan Szeider\\[4pt]
    \small LIACS, Leiden University & \small Algorithms and Complexity Group\\[-3pt]
    \small Leiden, Netherlands & \small TU Wien, Vienna, Austria\\
  \end{tabular}%
}
\date{}

\usepackage[textsize=footnotesize]{todonotes}

  \newenvironment{contriblist}{\begin{enumerate}[label=\textbf{(\arabic*)}, leftmargin=*]}{\end{enumerate}}
  \newcommand{\citem}{\item}

\makeatletter
\let\@dblfloat\@float
\let\end@dblfloat\end@float
\makeatother
\begin{document}

\maketitle
\thispagestyle{empty}

\begin{abstract}
Tree decision diagrams (TDDs), recently introduced by \citet{CapelliEtAl26}, are
representations of Boolean functions. They generalize OBDDs by replacing
variable orders with vtrees. The width of a TDD counts the distinct
subfunctions across each vtree cut. Counting subfunctions is blind to
linear structure. We introduce algebraic tree decision diagrams (ATDDs) by
replacing this count with the rank of the cut matrix. The maximum of these ranks over all cuts, the factor rank, is exactly the width of the ATDD. Every TDD
can be converted into an ATDD of no greater width. An ATDD is a tree tensor network in canonical form. ATDDs are exponentially more succinct than TDDs, and
quasi-polynomially more succinct than structured deterministic DNNFs. Our main result is that this gain in succinctness does not cost
canonicity. ATDDs are the first strict canonical form for a rank-based
representation over arbitrary vtrees. For each function and vtree there is exactly one
ATDD, over the rationals and without any
genericity assumption. The canonicity admits a
local characterization. Whether a tree tensor network is the ATDD of the
Boolean function it computes can be checked in polynomial time from its
tensors alone. Equivalence checking becomes a test of syntactic equality. The operation set includes
polynomial-time model counting, weighted model counting, and a test whether a tree tensor network represents a Boolean function.

\citet{CapelliEtAl26} asked for the complexity of changing the vtree of a TDD and suggested local vtree changes. We answer the question. After a rotation of the
vtree at one node, the new ATDD differs from the old one in two tensors only. Vtree change runs in output-polynomial time both for TDDs
and for ATDDs. 

We formalize and
verify our core results in Lean. We provide a prototype implementation
of TDD and ATDD compilers as a first empirical validation.
\end{abstract}

\section{Introduction}

Knowledge compilation is a domain of research on  tractable languages to represent functions. Tractable representations are designed to enable polynomial-time resolution of otherwise hard queries and polynomial-time algorithms for manipulating functions~\citep{Darwiche2002}. OBDDs (Ordered Binary Decision Diagrams) are a prime example of tractable representations for Boolean functions~\citep{Wegener00}. These decision diagrams where variables are read in a certain order admit polynomial-time algorithms for hard queries on Boolean formulas such as satisfiability testing or model counting, and  support standard Boolean operations in polynomial time, such as conjunction, negation, etc. For a given variable ordering every OBDD can easily be put in some minimal canonical form. Canonicity is particularly useful in practice as it reduces semantic equivalence to syntactic equality and thus makes it easy and rewarding to reuse known OBDDs in caching schemes. Even for tractable representations for which canonical forms are generally not minimal, like for SDDs (Sentential Decision Diagrams)~\citep{Darwiche11}, the practical benefits that come from using those canonical forms often win over the theoretical disadvantages~\citep{BroeckD15}.

Every representation of Boolean functions is exponentially large on some functions. A more succinct representation that keeps the same queries tractable handles more functions within a given size. We have in particular two applications in mind: probabilistic inference, which essentially boils down to weighted model counting, and equivalence checking for Boolean functions.

Tree decision diagrams (TDDs) are a tractable representation of Boolean functions with canonical forms, recently introduced by \citet{CapelliEtAl26}. TDDs generalize OBDDs by being based on vtrees instead of linear variable orders. A vtree is a rooted, ordered binary tree whose leaves are the variables. TDDs maintain all the features of OBDDs: a polynomial-time apply operation, tractable negation, and minimization to a canonical form, while being significantly more succinct. In this paper, we introduce a generalization of TDD which maintains all features listed before while being again more succinct. We call the new form ATDD for \emph{algebraic} TDD and present it as a tree tensor network, described below.

To present ATDD, we first recall that the size of a minimal TDD is polynomially connected to its width which, at a vtree node $t$, is the number of distinct subfunctions across the variable cut at~$t$, which separates the variables below $t$ from the remaining variables. The maximum of these counts is exactly the factor width of~\citet{BovaSzeider17}. A count of subfunctions ignores the linear dependencies among them. For instance, the hidden weighted bit function $\HWB_n$ on $n$ variables \citep{Bryant91} has exponentially many distinct subfunctions at some cut of every vtree, for infinitely many $n$, and thus cannot be represented with small TDDs \citep{CapelliEtAl26}, yet, as we show, these subfunctions always span a space of dimension at most $n+2$.

\begin{figure}[t]
\centering
\begin{subfigure}[c]{0.5\columnwidth}
\centering
\[
  M_{t}(f) =
  \begin{pNiceArray}{cccc}[first-row,last-col]
   {\scriptstyle{\bar x_3\bar x_4}}    &  {\scriptstyle{\bar x_3 x_4}}   & {\scriptstyle{x_3\bar x_4}}   &  {\scriptstyle{x_3x_4}}   \\
   0   & 0   & 1   & 1   & {\scriptstyle{\bar x_1\bar x_2}} \\
   0   & 1   & 1   & 0   & {\scriptstyle \bar x_1 x_2} \\
   1   & 1   & 0   & 0   & {\scriptstyle x_1\bar x_2} \\
   1   & 0   & 0   & 1   & {\scriptstyle x_1 x_2}\\
\end{pNiceArray}
\]
\end{subfigure}%
\begin{subfigure}[c]{0.45\columnwidth}
\centering
\begin{tikzpicture}[scale=1.0]
\node[circle,inner sep=2pt,draw,label={west:\scriptsize $r$}] (t) at (0,0) {};
\node[circle,inner sep=2pt,draw,label={west:\scriptsize $t$}] (l) at (-1,-1) {};
\node[circle,inner sep=2pt,draw,label={east:\scriptsize $s$}] (r) at (+1,-1) {};
\node[font=\scriptsize] (x1) at (-1.5,-2) {$x_1$};
\node[font=\scriptsize] (x2) at (-0.5,-2) {$x_2$};
\node[font=\scriptsize] (x3) at (+0.5,-2) {$x_3$};
\node[font=\scriptsize] (x4) at (+1.5,-2) {$x_4$};

\draw (r) -- (t) -- (l);
\draw (x1) -- (l) -- (x2);
\draw (x3) -- (r) -- (x4);
\end{tikzpicture}\end{subfigure}
\caption{The running example $f = \SUMBIT_2$, with the cut matrix $M_t(f)$
at the node $t$ of the balanced vtree on the right. At $t$, the TDD has
width $4$ and the ATDD has width $3$.}
\label{fig:running}
\end{figure}
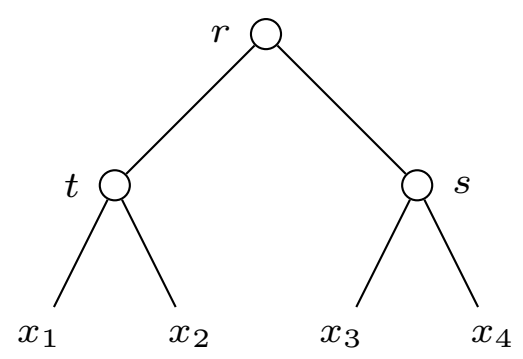

The running example of the paper is a function of four variables. We use this
to show how the count and the dimension differ. Let $f = \SUMBIT_2 = x_1
\oplus x_3 \oplus (x_2 \wedge x_4)$, the middle output bit of the sum of two
$2$-bit numbers $(x_1 x_2)_2 + (x_3 x_4)_2$, over the balanced vtree of
Figure~\ref{fig:running}, whose node $t$ has the variables $x_1, x_2$ below
it. The cut at $t$ is described by the \emph{cut matrix} $M_t(f)$ of
Figure~\ref{fig:running}. We index its rows by the assignments to $x_1, x_2$
and its columns by the assignments to $x_3, x_4$. The entries are the values
of $f$. All four rows are distinct and nonzero. So a minimal TDD has four
nodes at this cut. The last row, however, is the first row minus the second
plus the third. So the rows span a space of dimension $3$, and so do the columns. Consequently, the
ATDD introduced below has three vectors at this cut.

In an ATDD, we do not count the distinct subfunctions across a cut, but we measure the dimension of the space they span. This is the rank of the cut matrix, called the \emph{cut rank}. The \emph{factor rank} is the maximum of the cut ranks over all cuts of the vtree. As objects, ATDDs are tree tensor networks over vtrees. A tree tensor network for $f$ over $T$ has, at every node $t$ of $T$, a list of vectors indexed by the assignments to the variables below $t$. At its leaves, the vectors are given explicitly. At its internal nodes, the vectors are given in terms of a tensor of coefficients. Each tensor expresses these vectors as a linear combination of products of the vectors of the two children. The root has the single vector $f$. In the ATDD of $f$, the vectors at a node $t$ form a basis of the column space of the cut matrix. So the number is the cut rank at $t$. We fix the basis by the echelon form described in Section~\ref{sec:canon}. So the ATDD of $f$ is unique. The rank of a matrix is at most the number of its distinct nonzero rows. So the width of the ATDD is at most the width of the corresponding TDD, and it can be exponentially smaller. The main use of an ATDD is in knowledge compilation. A set of Boolean constraints is compiled once, and the ATDD then can be used for queries such as model counting, weighted model counting, or equivalence queries, and answers them in polynomial time.

Representations whose size is governed by matrix ranks have been studied before \citep{Waack97,BozapalidisAlexandrakis89,KieferMarusicWorrell15,Balle2015}, but none of them is canonical over vtrees. We call a representation \emph{strictly canonical} if, for a fixed vtree, every function has exactly one representation, with no freedom in the choice of a basis or of a scaling and no genericity assumption. The canonicity of a TDD is of this kind and follows directly from merging equal subfunctions into a minimal form. For rank-based representations, two networks can compute the same function with the same minimal dimensions, but still differ in every entry of their tensors. We resolve this by assigning to each cut a distinguished basis of its cut space, the column space of its cut matrix. We take the basis in its reduced row echelon form, the normal form of Gauss--Jordan elimination, which is unique. Such per-node bases must be compatible, so that the pivots of the basis at a node are products of the pivots at its children. The bases are compatible exactly when the coordinate order extends the product order of the coordinates of the children, in which $\tau_1 \times \tau_2$ precedes $\tau'_1 \times \tau'_2$ whenever $\tau_1 \le \tau'_1$ and $\tau_2 \le \tau'_2$ (Lemma~\ref{lem:compose}; the converse is proved in \suppl). The coordinates are the assignments below a node, and the coordinate order is the lexicographic order along the leaves of the vtree.

\paragraph{Contributions.}
\begin{contriblist}
\citem We define the ATDD of a Boolean function over a vtree and prove
that it is well defined and unique (Theorems~\ref{thm:sound}
and~\ref{thm:canonical}). ATDDs are the first strict canonical form for a
rank-based representation over arbitrary vtrees; concurrent work of \citet{Vilmart26} gives one for tensor trains, the case of a linear vtree (Section~\ref{sec:canon}). By Lemma~\ref{lem:compose}, the canonical tests, which are the pivots of the echelon bases above and form the lexicographically least set of assignments below a node whose subfunctions form a basis of the row space of its cut matrix, are
products of the canonical tests of its children.
\citem We show that a network that computes a Boolean function is the ATDD of that function
if and only if each of its tensors satisfies an echelon condition and a rank
condition (Theorem~\ref{thm:local}). This property can be checked in polynomial time
without reference to the represented function, and whether the function is
Boolean is itself decidable in polynomial time
(Theorem~\ref{thm:promise}). Every network computing a function $f \not\equiv 0$ can be minimized, and every network computing a Boolean function can be brought into ATDD form, in polynomial time (Proposition~\ref{prop:minimization},
Theorem~\ref{thm:makeCanonical}).
\citem We prove that the width of the ATDD is exactly the factor rank
(Proposition~\ref{prop:realization}). For $f \not\equiv 0$, the minimal dimension of a network
computing $f$ at a vtree node $t$ equals the cut rank $\rho_t$, so the ATDD
has size $O(n\rho^3)$ for factor rank $\rho$ and $n$ variables. TDDs embed
node by node as tree tensor networks with $0/1$ tensors, and the corresponding ATDDs are no wider
(Proposition~\ref{proposition:simulation}). ATDDs are exponentially
separated from TDDs via the hidden weighted bit function
(Theorem~\ref{thm:separation}), and quasi-polynomially separated from
structured deterministic DNNFs via a family due to \citet{VinallSmeeth24}.
\citem We show that many operations and tests run in polynomial time on
ATDDs: model counting and weighted model counting, conditioning,
conjunction, disjunction, negation and single-variable forgetting
(Theorems~\ref{thm:modelCounting} to~\ref{thm:forgetting}).
\citem We show that after a rotation of the vtree at one node, the ATDD
of the new vtree differs from the old one in two tensors only, which we compute
from the two old ones (Theorem~\ref{thm:rotation}). An ATDD can
be converted from one vtree to another in time polynomial in the input and
the output (Theorem~\ref{thm:change}); the same holds for TDDs, which
answers the question of \citet{CapelliEtAl26}.

\citem We formalize our core results in Lean~4, which allows
mechanical checking of the results. We also test the algorithms by
exact-arithmetic differential testing. We provide prototype compilers for TDDs
and ATDDs. These compilers allow us to quantify the differences in a
preliminary empirical study.
\end{contriblist}

\begin{figure}[t]
\centering
\includegraphics[width=\wlandscape\columnwidth]{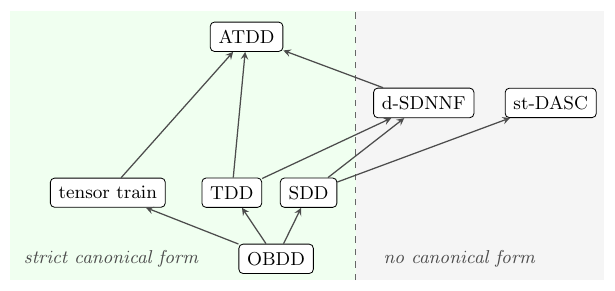}
\vspace{-10pt}
\caption{Relevant representations by succinctness and canonicity; an
arrow from $x$ to $y$ means that $y$ polynomially simulates $x$.}
\label{fig:landscape}
\end{figure}

\paragraph{Related representations.}
Figure~\ref{fig:landscape} compares ATDDs with
other representations in terms of their canonicity and succinctness. Weighted
tree automata have minimal representations based on Hankel ranks. But these are
unique only up to a change of basis
\citep{BozapalidisAlexandrakis89,KieferMarusicWorrell15}. The singular-value
refinements \citep{Balle2015,Rabusseau2016} are unique only for distinct
singular values. Tree tensor networks
with minimal bond dimensions are the hierarchical Tucker format
\citep{HackbuschKuhn09,Grasedyck10}. None of them has a strict canonical form
or a Boolean layer, a dedicated treatment of $0/1$-valued
functions. The matrix product states of \citet{VinkhuijzenCoopmansLaarman26}, a representation designed for quantum states, are also called tensor trains and have no Boolean layer. For tensor trains, however, \citet{Vilmart26} gives a strict canonical form in concurrent work, by echelon forms instead of singular values.

Two recently proposed representations are related to our setting: tensor trains
over linear orders with entries in $\{-1,0,1\}$ \citep{OnakaEtAl25tt}, and
vtree-structured and-sum circuits with signed edges (st-DASCs)
\citep{OnakaEtAl25dasc}. The st-DASCs reach the same quasi-polynomial gap over
d-SDNNF that we do. However, neither paper gives a canonical form. In
st-DASCs, the Boolean promise, the requirement that the computed
function is $0/1$-valued, is imposed syntactically, without a test.

Finally, \citet{QuistEtAl26} translate between tensor networks and
structured-decomposable arithmetic circuits over a semiring; our d-SDNNF simulation (Proposition~\ref{proposition:dsdnnf}) is a direct proof of a rank bound that also follows from their translation of circuits into tensor networks. We add the Boolean
layer, the factor rank, and the strict canonical form; ATDDs are thus
the only representation among the most succinct ones that is canonical.

\paragraph{Additional material.}
The proofs of statements labeled with a star ($\star$) are in the Zenodo repository~\citep{zenodo}. All other proofs are in this paper; those of Theorems~\ref{thm:local}, \ref{thm:makeCanonical} and~\ref{thm:conditioning} each use one result proved there. In Part III of the repository we also bound the cut rank of a CNF at a cut of a given vtree by $(m+1)^k$, where $m$ is the number of clauses crossing the cut and $k$ is the smaller of the induced-matching numbers of the two bipartite graphs between the crossing clauses and the variables of either side; that result is not included in this version of the paper. We follow a general trend in mathematics, where
results are increasingly verified by machine-checkable proofs \citep{Tao25}. All
statements with a full Lean proof are labeled with a dagger ($\dagger$). Running times and bit sizes are not machine-checked, because the formalization has no cost model and bounds only the entries of the canonical form, not the numbers written by the algorithms. The Lean proofs use only the three standard axioms of Lean (propositional extensionality, quotient soundness and choice) and no \texttt{native\_decide}, with which the kernel would accept results of compiled code that it does not check. The Lean sources are part of
the Zenodo repository, with a statement-by-statement map
to the paper. There we also work out one example in full.

\section{Preliminaries}\label{sec:prelim}

Let $X$ be a finite set of propositional variables. We write $2^A$ for the set
of assignments $\tau: A \rightarrow \{0,1\}$, and $\tau \times \sigma \in 2^{A
\cup B}$ for the combination of $\tau \in 2^A$ and $\sigma \in 2^B$ when $A \cap
B = \emptyset$; $[k] = \{1, \dots, k\}$. A Boolean function over $X$ is a mapping
$f: 2^X \rightarrow \{0,1\}$, identified with a $0/1$-valued element of~$\Q^{2^X}$. For $\tau \in 2^Y$ with $Y \subseteq X$, the subfunction
$f[\tau]$ maps $\sigma \in 2^{X \setminus Y}$ to $f(\tau \times \sigma)$. For
disjoint~$A, B$ and~$u \in \Q^{2^A}$, $v \in \Q^{2^B}$, the product $u \otimes v
\in \Q^{2^{A \cup B}}$ is given by $(u \otimes v)(\tau \times \sigma) =
u(\tau)\, v(\sigma)$.

A \emph{vtree} $T$ over $X$ is a rooted, ordered, binary tree whose leaves
correspond bijectively to~$X$. $X_t$ denotes the variables below node~$t$. For a
Boolean function $f$ and a node $t$, the \emph{cut matrix} $M_t(f)$ has rows
indexed by $2^{X_t}$ and columns by $2^{X \setminus X_t}$, with entry $f(\tau
\times \sigma)$. The row of $\tau$ is the subfunction $f[\tau]$. We write
$S_t(f)$ for the number of distinct nonzero rows, the \emph{subfunction count}
at~$t$; $\fw(f,T) = \max_t S_t(f)$ is the \emph{factor width}
\citep{BovaSzeider17,CapelliEtAl26}, and, for~$f \not\equiv 0$, the width at $t$ of the minimal TDD, defined in Section~\ref{sec:width},
equals $S_t(f)$. We define
\[
  \rho_t(f) = \rk M_t(f)
  \qquad\text{and}\qquad
  \fr(f,T) = \max_{t \in T} \rho_t(f),
\]
the \emph{cut rank} $\rho_t(f)$ and the \emph{factor rank} $\fr(f,T)$ of $f$ with
respect to $T$, over $\Q$. The column space of~$M_t(f)$, spanned by the conditioned functions $f[\sigma]$, is the \emph{cut
space} $V_t(f) \subseteq \Q^{2^{X_t}}$, of dimension~$\rho_t(f)$. Formally, for each $\sigma \in 2^{X\setminus X_t}$, we see $f[\sigma]$ as a vector of $0$ and $1$ of size $2^{|X_t|}$ and define
$$
V_t(f) = \text{span}_{\Q}\{f[\sigma] : \sigma \in 2^{X\setminus X_t}\}.
$$
A standard result is that the cut space $V_t$ is nested in the tensor product of the cut spaces of $t$'s children. The proof is provided in Appendix~\ref{app:nested} for convenience.

\begin{restatable}[$\dagger$]{proposition}{lemnested}\label{proposition:nestedness}
Let $f \not\equiv 0$ be a Boolean function over $X$ and $T$ be a vtree over $X$. For $t$ an internal node of $T$ with children $t_1$ and $t_2$, we have $V_t \subseteq V_{t_1} \otimes V_{t_2}$. 
\end{restatable}

\begin{example}\label{ex:sumbit}
The running example is $\SUMBIT_2$ of the introduction over the balanced
vtree of Figure~\ref{fig:running}. Figures~\ref{fig:tdd} and~\ref{fig:atdd}
use the same vtree. At the node $t$ with $X_{t} = \{x_1, x_2\}$, the cut
matrix~$M_t(f)$ is the matrix of Figure~\ref{fig:running}. Its four rows are
distinct and nonzero, so $S_{t} = 4$, but $f[x_1 x_2] = f[\bar x_1\bar x_2] -
f[\bar x_1 x_2] + f[x_1\bar x_2]$, so $\rho_{t} = 3$.
\end{example}

\begin{definition}\label{def:ttn}
All tensors and matrices are over $\Q$. A \emph{tree tensor network} (for short: \emph{TTN}) respecting a vtree $T$
over $X$ is a triple $A = (d, u, W)$. 
\begin{itemize}
\item[•] $d$ assigns a \emph{bond dimension} $d(t) \ge 1$ to every
node $t$, with $d(r) = 1$ at the root.
\item[•] $W$ assigns a $d(t) \times d(t_1) \times d(t_2)$ \emph{transfer tensor} $W^t$ to every internal $t$ with children $t_1, t_2$. $\bar W^t$, the \emph{parent unfolding} of $W^t$, is the $d(t) \times d(t_1)d(t_2)$ matrix defined by $\bar W^t_{i,(j-1)d(t_2)+k} = W^t_{ijk}$, so that the columns are indexed by the pairs $(j,k)$ in lexicographic order.
\item[•] $u$ assigns a $d(t) \times 2^{\{x\}}$ \emph{leaf matrix} $u^t$ to every leaf $t$ labeled $x$. The $(i,\tau) \in [d(t)]\times 2^{\{x\}}$ entry of $u^t$ is denoted by $u^t_i(\tau)$.
\end{itemize} 
$u^t$ is also defined for internal nodes as a $d(t) \times 2^{X_t}$ matrix such that
\begin{equation}\label{eq:evaluate_ttn}
  u^t_i(\tau) \;=\; \sum_{j \in [d(t_1)]} \sum_{k \in [d(t_2)]}
      W^t_{ijk}\, u^{t_1}_j(\tau_1) u^{t_2}_k(\tau_2), \quad \text{i.e.},\quad u^t = \bar{W}^t (u^{t_1} \otimes u^{t_2})
\end{equation}
where $\tau_1$ and $\tau_2$ are the restrictions of $\tau$ to $X_{t_1}$ and $X_{t_2}$, respectively, and $(u^{t_1} \otimes u^{t_2})$ is the $d(t_1)d(t_2) \times 2^{X_t}$ matrix defined by $(u^{t_1} \otimes u^{t_2})_{(j-1)d(t_2)+k}(\tau) = u^{t_1}_j(\tau_1) u^{t_2}_k(\tau_2)$.

The TTN $A$ \emph{computes} $f_A = u^{r}_1$; its \emph{width} is $\max_t d(t)$ and its size is the number of entries, $O(n w^3)$ for width $w$.
\end{definition}

Summing a product of tensors over their shared indices is called \emph{contraction}; computing $f_A(\tau)$ by \eqref{eq:evaluate_ttn} contracts the whole network. A TTN may compute any function from $2^X$ to $\Q$; the requirement that $f_A$ is $0/1$-valued is the \emph{Boolean promise}, which constrains only the root and is decidable in polynomial time (Theorem~\ref{thm:promise}). The new object of this paper is the \emph{ATDD} of a Boolean function, the canonical TTN of the function (Section~\ref{sec:canon}). In this sense, our main result is that \emph{tree tensor networks, in the Boolean setting over a vtree, can be made canonical}.

\section{Succinctness of TTNs}

\subsection{Factor Rank Is the Smallest TTN Width}\label{sec:width}

\begin{figure}[t]
\centering
\begin{subfigure}{0.5\textwidth}
\centering
\begin{tikzpicture}
\def\x{10pt};
\def\y{1.75};
\node[draw,rectangle,inner sep=5pt] (n0) at (0,0) {};
\node[draw,rectangle,inner sep=5pt] (n1) at (\x,0) {};
\node[draw,rectangle,inner sep=5pt] (n2) at (2*\x,0) {};
\node[draw,rectangle,inner sep=5pt] (n3) at (3*\x,0) {};
\node[draw,rectangle,inner sep=5pt] (n4) at (4*\x,0) {};
\node[draw,rectangle,inner sep=5pt] (n5) at (5*\x,0) {};
\node[draw,rectangle,inner sep=5pt] (n6) at (6*\x,0) {};
\node[draw,rectangle,inner sep=5pt] (n7) at (7*\x,0) {};

\node[draw,rectangle,inner sep=5pt] (l0) at (-5*\x,-\y) {};
\node[draw,rectangle,inner sep=5pt] (l1) at (-3*\x,-\y) {};
\node[draw,rectangle,inner sep=5pt] (l2) at (-1*\x,-\y) {};
\node[draw,rectangle,inner sep=5pt] (l3) at (1*\x,-\y) {};

\node[draw,rectangle,inner sep=5pt] (r0) at (6*\x,-\y) {};
\node[draw,rectangle,inner sep=5pt] (r1) at (8*\x,-\y) {};
\node[draw,rectangle,inner sep=5pt] (r2) at (10*\x,-\y) {};
\node[draw,rectangle,inner sep=5pt] (r3) at (12*\x,-\y) {};

\node (b1) at (-5*\x,-2*\y) {$\bar x_1$};
\node (x1) at (-3*\x,-2*\y) {$x_1$};
\node (b2) at (-1*\x,-2*\y) {$\bar x_2$};
\node (x2) at (1*\x,-2*\y) {$x_2$};

\node (b3) at (6*\x,-2*\y) {$\bar x_3$};
\node (x3) at (8*\x,-2*\y) {$x_3$};
\node (b4) at (10*\x,-2*\y) {$\bar x_4$};
\node (x4) at (12*\x,-2*\y) {$x_4$};

\draw[-latex] (n0.south) to [out=-90,in=90] (l0.north);
\draw[-latex] (n0.south) to [out=-45] (r2);

\draw[-latex] (n1.south) to [out=-90,in=90] (l0.north);
\draw[-latex] (n1.south) to [out=-45] (r3);

\draw[-latex] (n2.south) to [out=-90,in=90] (l1.north);
\draw[-latex] (n2.south) to [out=-45] (r1);

\draw[-latex] (n3.south) to [out=-90,in=90] (l1.north);
\draw[-latex] (n3.south) to [out=-45] (r2);

\draw[-latex] (n4.south) to [out=-90,in=90] (l2.north);
\draw[-latex] (n4.south) to [out=-90] (r0);

\draw[-latex] (n5.south) to [out=-90,in=90] (l2.north);
\draw[-latex] (n5.south) to [out=-90] (r1);

\draw[-latex] (n6.south) to [out=-90,in=90] (l3.north);
\draw[-latex] (n6.south) to [out=-90] (r0);

\draw[-latex] (n7.south) to [out=-90,in=90] (l3.north);
\draw[-latex] (n7.south) to [out=-90] (r3);

\draw[-latex] (l0.south) -- (b1.north);
\draw[-latex] (l0.south) -- (b2.north);

\draw[-latex] (l1.south) -- (b1.north);
\draw[-latex] (l1.south) -- (x2.north);

\draw[-latex] (l2.south) -- (x1.north);
\draw[-latex] (l2.south) -- (b2.north);

\draw[-latex] (l3.south) -- (x1.north);
\draw[-latex] (l3.south) -- (x2.north);

\draw[-latex] (r0.south) -- (b3.north);
\draw[-latex] (r0.south) -- (b4.north);

\draw[-latex] (r1.south) -- (b3.north);
\draw[-latex] (r1.south) -- (x4.north);

\draw[-latex] (r2.south) -- (x3.north);
\draw[-latex] (r2.south) -- (b4.north);

\draw[-latex] (r3.south) -- (x3.north);
\draw[-latex] (r3.south) -- (x4.north);

\end{tikzpicture}
\end{subfigure}\begin{subfigure}{0.5\textwidth}
\centering
\begin{tikzpicture}[scale=0.75, every node/.style={scale=0.75}]
  \atddtable{r}{(0,4.0)}{
    r \& 11\&12\&13\&14\&21\&22\&23\&24\&31\&32\&33\&34\&41\&42\&43\&44 \\
    1 \& 0\&0\&1\&1\&0\&1\&1\&0\&1\&1\&0\&0\&1\&0\&0\&1 \\ }

  \atddtable{t}{(-2.6,2.0)}{
    t \& 11\&12\&21\&22 \\ 1\&1\&0\&0\&0 \\ 2\&0\&1\&0\&0 \\ 3\&0\&0\&1\&0 \\ 4\&0\&0\&0\&1 \\ }
  \atddtable{s}{(2.6,2.0)}{
    s \& 11\&12\&21\&22 \\ 1\&1\&0\&0\&0 \\ 2\&0\&1\&0\&0 \\ 3\&0\&0\&1\&0 \\ 4\&0\&0\&0\&1 \\ }

  \atddtable{x1}{(-3.35,-0.45)}{ x_1 \& 0\&1 \\ 1\&1\&0 \\ 2\&0\&1 \\ }
  \atddtable{x2}{(-1.55,-0.45)}{ x_2 \& 0\&1 \\ 1\&1\&0 \\ 2\&0\&1 \\ }
  \atddtable{x3}{( 1.55,-0.45)}{ x_3 \& 0\&1 \\ 1\&1\&0 \\ 2\&0\&1 \\ }
  \atddtable{x4}{( 3.35,-0.45)}{ x_4 \& 0\&1 \\ 1\&1\&0 \\ 2\&0\&1 \\ }

  \atddedge{t}{r}{4}  \atddedge{s}{r}{4}
  \atddedge{x1}{t}{2} \atddedge{x2}{t}{2}
  \atddedge{x3}{s}{2} \atddedge{x4}{s}{2}
\end{tikzpicture}
\end{subfigure}
\caption{The minimal TDD for $\SUMBIT_2$ and its TTN translation: four nodes at $t$ and $s$, one per distinct subfunction; the single root node is drawn as its eight input pairs (pairs of nodes at $t$ and $s$), each pointing at its two nodes. The count $S_t=4$ exceeds the cut
rank $\rho_t=3$.}
\label{fig:tdd}
\end{figure}
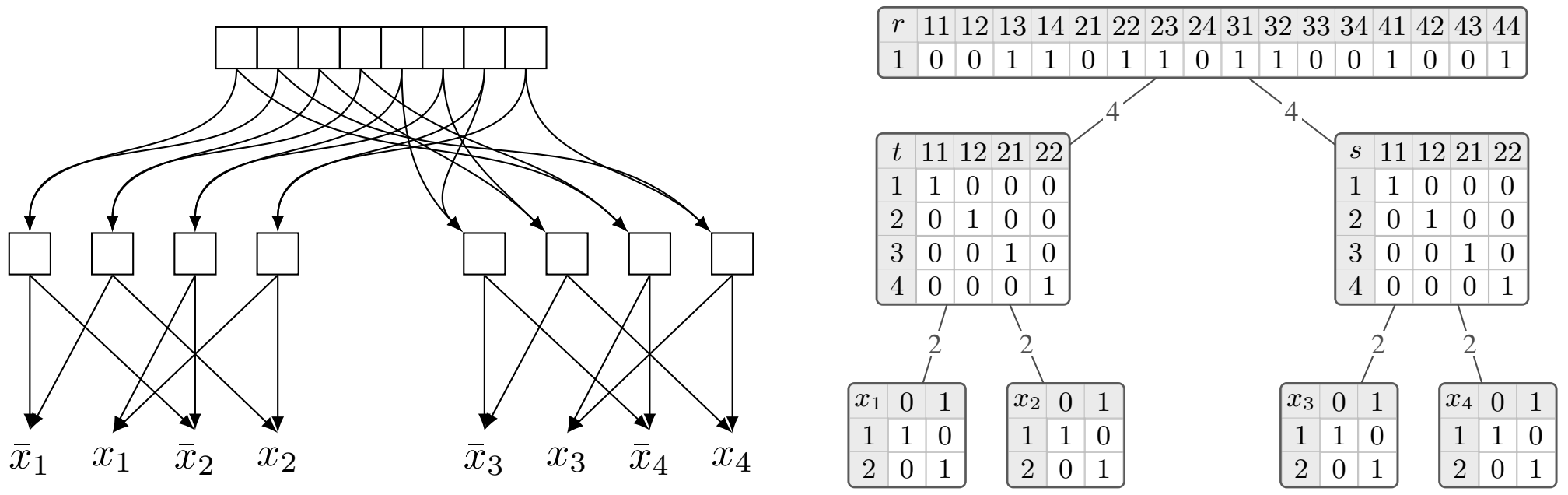

In this section, we show that the factor rank of a Boolean function $f \not\equiv 0$ is exactly the least width of a TTN computing $f$. A TTN in which the dimension equals the cut rank for every~$t$, i.e., $d(t) = \rho_t(f)$, is called a \emph{minimal TTN}.

\begin{proposition}[$\dagger$]\label{prop:realization}
Every TTN computing $f$ has $d(t) \ge \rho_t(f)$ at every node. For every Boolean function~$f \not\equiv 0$ and vtree, a minimal TTN exists, with size $O(|X| \cdot \fr(f,T)^3)$.
\end{proposition}
\begin{proof}
The lower bound follows from (\ref{eq:min-split}) below, whose derivation does not use this proposition: it writes $M_t(f)$ as the product $(u^t)^{\top} \bar u^t$ of matrices with $d(t)$ rows, so $\rho_t(f) = \rk M_t(f) \le d(t)$.

For the existence, since $f \not\equiv 0$, every cut space $V_t$ has dimension $\rho_t(f) \ge 1$. For every node $t$, choose a basis $b^t_1, \dots, b^t_{\rho_t(f)}$ of $V_t$, with $b^r_1 = f$ at the root $r$, and set $d(t) = \rho_t(f)$. If~$t$ is a leaf, set $u^t_i = b^t_i$. If $t$ is an internal node with children $t_1$ and $t_2$, then $V_t \subseteq V_{t_1} \otimes V_{t_2}$ by Proposition~\ref{proposition:nestedness}, and the products $b^{t_1}_j \otimes b^{t_2}_k$ form a basis of $V_{t_1} \otimes V_{t_2}$ by Lemma~\ref{lem:app-basis}, so there are coefficients $W^t_{ijk}$ such that
\begin{align*}
b^t_i = \sum_{j \in [d(t_1)]} \sum_{k \in [d(t_2)]} W^t_{ijk}\, b^{t_1}_j \otimes b^{t_2}_k.
\end{align*}
By induction and (\ref{eq:evaluate_ttn}), $u^t_i = b^t_i$ at every node $t$, so $u^r_1 = f$ and this TTN computes $f$. It has $d(t) = \rho_t(f) \le \fr(f,T)$ at every node, so each of its $2|X| - 1$ nodes stores $O(\fr(f,T)^3)$ entries, which gives the size $O(|X| \cdot \fr(f,T)^3)$.
\end{proof}

Consequently, for $f \not\equiv 0$, the factor rank is the least width of a network computing $f$. The lower bound is given by the factorization $M_t(f) = (u^t)^{\top} \bar u^t$ of \eqref{eq:min-split}, which exhibits $M_t(f)$ as a sum of $d(t)$ rank-one matrices. For the upper bound, we choose a basis of each cut space and use that the cut spaces are nested, $V_t \subseteq V_{t_1} \otimes V_{t_2}$. 

Finding a vtree that minimizes the factor rank is a hard task but, if the vtree is fixed, then bringing a TTN to an equivalent minimal form is tractable.

\begin{proposition}\label{prop:minimization}
Given a TTN computing $f$ respecting the vtree $T$, one decides whether $f \equiv 0$ and, if not, constructs a minimal TTN for $f$ respecting $T$, in polynomial time.
\end{proposition}

We describe the minimization procedure, and then prove Proposition~\ref{prop:minimization}. We assume $f \not\equiv 0$; the procedure detects $f \equiv 0$ (some matrix it factors is zero, or, if $T$ is a single leaf, the leaf matrix is zero), and then returns the ATDD of the $0$ function (Definition~\ref{def:canonical}). Besides the parent unfolding $\bar W^t$ of Definition~\ref{def:ttn}, we use the two \emph{child unfoldings} of a transfer tensor: $W^t_{\langle t_1 \rangle} \in \Q^{d(t_1) \times d(t)d(t_2)}$, whose row $j$ is $(W^t_{ijk})_{(i,k)}$, and $W^t_{\langle t_2 \rangle} \in \Q^{d(t_2) \times d(t)d(t_1)}$, whose row $k$ is $(W^t_{ijk})_{(i,j)}$, with the pairs $(i,k)$, resp.\ $(i,j)$, in lexicographic order.

The matrix $u^t \in \Q^{d(t) \times 2^{X_t}}$ of Definition~\ref{def:ttn} describes the network below $t$. Its rows $u^t_i$ are the \emph{bond vectors} at $t$.
The network above $t$ is described by a matrix $\bar u^t \in \Q^{d(t) \times 2^{X \setminus X_t}}$, whose rows $\bar u^t_i$ are the \emph{context vectors} at $t$. It is defined top-down by $\bar u^r = (1)$ at the root (a $1 \times 1$ matrix, since $X \setminus X_r = \emptyset$) and, for an internal $t$ with children $t_1, t_2$,
\begin{equation}\label{eq:min-above}
  \bar u^{t_1} \;=\; W^t_{\langle t_1 \rangle}\, (\bar u^t \otimes u^{t_2}),
  \qquad
  \bar u^{t_2} \;=\; W^t_{\langle t_2 \rangle}\, (\bar u^t \otimes u^{t_1}),
\end{equation}
where $\bar u^t \otimes u^{t_2}$ is the $d(t)d(t_2) \times 2^{X \setminus
X_{t_1}}$ matrix whose row $(i,k)$ is $\bar u^t_i \otimes u^{t_2}_k$, with the pairs $(i,k)$ in lexicographic order, and
symmetrically for $\bar u^t \otimes u^{t_1}$. At every node $t$,
\begin{equation}\label{eq:min-split}
  M_t(f) \;=\; (u^t)^{\top} \bar u^t,
  \qquad\text{that is,}\qquad
  f(\tau \times \sigma) \;=\; \sum_{i \in [d(t)]} u^t_i(\tau)\, \bar u^t_i(\sigma)
\end{equation}
for all $\tau \in 2^{X_t}$ and $\sigma \in 2^{X \setminus X_t}$. At the
root this is $f = u^r_1$. If it holds at $t$, substituting $u^t_i(\tau) =
\sum_{j,k} W^t_{ijk}\, u^{t_1}_j(\tau_1)\, u^{t_2}_k(\tau_2)$ from
\eqref{eq:evaluate_ttn} and collecting the terms by $j$ gives
$\sum_j u^{t_1}_j(\tau_1)\, \bar u^{t_1}_j(\tau_2 \times \sigma)$, so
\eqref{eq:min-split} holds at $t_1$, and symmetrically at $t_2$.

\paragraph{Rank criterion.}
By \eqref{eq:min-split}, $\rho_t(f) = \rk\bigl((u^t)^{\top} \bar u^t\bigr) \le
d(t)$, with equality if and only if both~$u^t$ and~$\bar u^t$ have full row
rank. Indeed, equality forces $\rk u^t = \rk \bar u^t = d(t)$; conversely, if~$u^t$ has full row rank, then $(u^t)^{\top}$ has a left inverse, so
$\rk\bigl((u^t)^{\top} \bar u^t\bigr) = \rk \bar u^t$. Hence $A$ is minimal if and only if~$u^t$ and $\bar u^t$ have full row rank at every node $t$.

We use two facts about full row rank. First, if $u$ and $u'$ have full row
rank and live on disjoint sets of variables, then the matrix with rows $u_j
\otimes u'_k$ has full row rank (Lemma~\ref{lem:app-basis}). Second, if $M$
and $N$ have full row rank, then so does $MN$: $x^{\top} M N = 0$ gives
$x^{\top} M = 0$, since the rows of $N$ are independent, and then $x = 0$.

\paragraph{Minimization procedure.}
Pass~1 makes every $u^t$ of full row rank, bottom-up; Pass~2 then makes every
$\bar u^t$ of full row rank, top-down, while keeping the $u^t$ of full row rank.
\begin{enumerate}
\item[(1)] \emph{Pass~1, bottom-up.} Process the non-root nodes in
  post-order. Consider a node $t$ with parent $p$; at a leaf, $\bar W^t$ denotes the leaf matrix $u^t$.
  \begin{itemize}
  \item Compute $r = \rk \bar W^t$ and factor $\bar W^t = P\, \hat W^t$, where $P \in \Q^{d(t) \times r}$ has full column rank and $\hat W^t$ has $r$ rows.
  \item If $t$ is the left child of $p$ and $s$ its sibling, set
  \[
    W'^p_{hak} \;=\; \sum_{i \in [d(t)]} W^p_{hik}\, P_{ia}
    \qquad (h \in [d(p)],\ a \in [r],\ k \in [d(s)]),
  \]
  and symmetrically along the third index if $t$ is the right child.
  \item Replace $\bar W^t$ by $\hat W^t$, folded back, so that $d(t)$ becomes $r$.
  \end{itemize} 
\item[(2)] \emph{Pass~2, top-down.} Process the internal nodes in
  pre-order. At node $t$, for each child $c$ consider $W^t_{\langle c \rangle} \in \Q^{d(c)
  \times d(t)d(c')}$, with $c'$ the sibling of $c$.
  \begin{itemize}
  \item Compute $r = \rk W^t_{\langle c \rangle}$ and factor $W^t_{\langle c \rangle} = Q^{\top} \hat W^t_{\langle c \rangle}$, where $Q \in \Q^{r \times d(c)}$ has full row rank and $\hat W^t_{\langle c \rangle}$ has $r$ rows. 
  \item Replace $W^t$ by $\hat W^t_{\langle c \rangle}$, folded back, so that $d(c)$ becomes $r$. 
  \item Replace $\bar W^c$ by $Q\, \bar W^c$ (at a leaf, $u^c$ by $Q\, u^c$).
  \end{itemize}
\end{enumerate}
In both passes $r \ge 1$. In Pass~1, $\bar W^t = 0$ would give $u^t = 0$
by \eqref{eq:evaluate_ttn}, and in Pass~2, $W^t_{\langle c \rangle} = 0$ would give $\bar u^c = 0$ by
\eqref{eq:min-above}; in both cases $f = 0$ by \eqref{eq:min-split}. The
root is not processed in Pass~1: $\bar W^r$ is a single row, which is not
zero for the same reason. Conversely, if $f \equiv 0$ and no matrix factored in Pass~1 is zero, then after Pass~1 the matrices $u^{r_1}$ and $u^{r_2}$ at the children of the root have full row rank (proof below), so $0 = u^r = \bar W^r (u^{r_1} \otimes u^{r_2})$ gives $\bar W^r = 0$, and the first step of Pass~2 factors a zero matrix. Since $\hat W^t$ and $\hat W^t_{\langle c \rangle}$ have rank $r$ and
$r$ rows, both have full row rank.

\begin{proof}[Proof of Proposition~\ref{prop:minimization}]

Let $t$ be processed in Pass~1. By \eqref{eq:evaluate_ttn}, $u^t = \bar
W^t\, (u^{t_1} \otimes u^{t_2}) = P\, \hat u^t$ with $\hat u^t = \hat W^t\,
(u^{t_1} \otimes u^{t_2})$ (at a leaf, $u^t = P\, \hat u^t$ with $\hat u^t =
\hat W^t$). After the step, the matrix at $t$ is $\hat u^t$, and the
matrix at $p$ is unchanged, since
\[
  \sum_{a,k} W'^p_{hak}\, \hat u^t_a \otimes u^s_k
  \;=\; \sum_{i,k} W^p_{hik} \Bigl(\sum_a P_{ia}\, \hat u^t_a\Bigr) \otimes
  u^s_k
  \;=\; \sum_{i,k} W^p_{hik}\, u^t_i \otimes u^s_k \;=\; u^p_h .
\]
So $f$ is unchanged, and so is $u^v$ for every node $v \ne t$: the nodes
below $t$ and the sibling's subtree are untouched, and the matrices at the
ancestors of $p$ depend on the subtree of $p$ only through $u^p$.

When $t$ is processed, its children have been processed before, so $u^{t_1}$ and $u^{t_2}$ have full row rank, and then so does $u^{t_1} \otimes u^{t_2}$. Hence $\hat u^t = \hat W^t\, (u^{t_1} \otimes u^{t_2})$ has full row rank; at a leaf, $\hat u^t = \hat W^t$ has full row rank directly. The later steps do not change $u^t$, by the previous paragraph. At the root, $u^r = f \ne 0$ is a single nonzero row.

One step of Pass 2. Let the step at $t$ concern its child $c = t_1$ (the case $c = t_2$ is
symmetric), and write $B = Q^{\top} \hat B$ entrywise as $W^t_{ijk} =
\sum_{a \in [r]} Q_{aj}\, \hat B_{a,(i,k)}$. After the step, $W'^t_{iak} =
\hat B_{a,(i,k)}$ and the matrix at $t_1$ is $Q\, u^{t_1}$, so the matrix at
$t$ is unchanged:
\[
  \sum_{a,k} W'^t_{iak}\, (Q u^{t_1})_a \otimes u^{t_2}_k
  \;=\; \sum_{j,k} \Bigl(\sum_a Q_{aj}\, \hat B_{a,(i,k)}\Bigr)\,
  u^{t_1}_j \otimes u^{t_2}_k
  \;=\; u^t_i .
\]
Hence $f$ is unchanged, and so is $u^v$ for every node $v \ne t_1$. The
same expansion, with $\bar u^t_i$ in place of~$u^{t_2}_k$, shows that $\bar u^{t_2}$ is
unchanged, and by \eqref{eq:min-above} the new matrix at $t_1$ is
\[
  \hat{\bar{u}}^{t_1} \;=\; \hat B\, (\bar u^t \otimes u^{t_2}),
  \qquad\text{with}\qquad \bar u^{t_1} = Q^{\top} \hat{\bar{u}}^{t_1}.
\]
The matrices $\bar u^v$ are computed top-down by \eqref{eq:min-above}, from the
transfer tensors at the strict ancestors of $v$ and from the matrices $u$ at
the siblings of $v$ and of its ancestors. The step changes only the tensors
at~$t$ and $t_1$ and the matrix $u^{t_1}$, so $\bar u^v$ is unchanged for every
node $v$ outside the subtree of $t_1$; at $t_1$ it becomes $\hat{\bar{u}}^{t_1}$.
Finally, if $u^{t_1}$ has full row rank, then so does $Q\, u^{t_1}$ (second
fact).

By the previous paragraph and Pass~1, every $u^v$ has full row rank
throughout Pass~2. We show that $\bar u^t$ has full row rank once the step at the
parent $p$ of $t$ concerning $t$ has been made, and keeps it until the end of
the pass; at the root, $\bar u^r = (1)$. Say $t = p_1$. Before the step, $\bar u^p$
has full row rank by induction, since the step at the parent of $p$ concerning
$p$ comes before in pre-order, and $u^{p_2}$ has full row rank. So $\bar u^p
\otimes u^{p_2}$ has full row rank (first fact), and so does $\hat{\bar{u}}^{t} =
\hat B\, (\bar u^p \otimes u^{p_2})$ (second fact). The case $t = p_2$ is the
same, with $u^{p_1}$, whose full row rank is kept by the first step at $p$.
After that, $\bar u^t$ changes only through a step concerning an ancestor of $t$
(previous paragraph), and in pre-order all such steps were made before.

\paragraph{Minimality.}
After both passes, $u^t$ and $\bar u^t$ have full row rank at every node $t$, so
$d(t) = \rho_t(f)$ at every node by the rank criterion. The resulting TTN
respects $T$ and still computes $f$, so it is a minimal TTN for $f$.

\paragraph{Cost.}
Let $w$ be the width of $A$; no step increases a dimension. Each step
factors a matrix with at most~$w$ rows and at most $\max(w^2, 2)$ columns, in $O(w^4)$
arithmetic operations by Gaussian elimination, and absorbs a $w \times w$
matrix into a tensor with at most $w^3$ entries, in $O(w^4)$ arithmetic
operations. Pass~1 makes one step per non-root node and Pass~2 two steps per
internal node, so the procedure runs with $O(|X| \cdot w^4)$ arithmetic
operations.

The bit sizes stay polynomial for the following choice of the factorizations.
In Pass~1, $P$ is made of the pivot columns of $\bar W^t$, and $\hat W^t$ of
the nonzero rows of the reduced row echelon form of $\bar W^t$. In Pass~2,
$\hat B$ is made of the rows of $B$ indexed by the lexicographically first
set~$S$ of rows forming a basis of the row space, and $Q^{\top}$ holds the
coordinates of the rows of $B$ in that basis, so that the columns of~$Q$
indexed by $S$ form an identity matrix.

\emph{Pass~1.} Every entry of $P$ is an entry of $\bar W^t$ at the time $t$
is processed, and $\bar W^t$ at that time is the input tensor multiplied by
the matrices $P$ of the children. Unfolding this recursion, every such entry
is a polynomial in the entries of $A$ of degree at most $2|X|$ with at most
$w^{2|X|}$ monomials, whose denominators divide the product of the
denominators of all entries of $A$; so it has bit size polynomial in the bit size
of $A$. The entries of $\hat W^t$ are ratios of minors of $\bar W^t$, hence
also of polynomial bit size. Consequently, the TTN $A_1$ obtained after
Pass~1 has entries of polynomial bit size, and so do the entries of its
matrices $u^t$ and of $f$, by the same counting.

\emph{Pass~2.} Each step at $t$ concerning $c$ keeps a subset of the slices
of $W^t$ and multiplies the tensor at~$c$ by $Q$; it creates no other number.
Each tensor other than the root tensor is multiplied by exactly one matrix $Q$, when its parent is
processed. It remains to bound the entries of $Q$, and they are determined by
$A_1$ and $f$ alone. Indeed, when the step is made, $u^c$ is still the matrix
of $A_1$, and $\bar u^c$ is the unique solution of $(u^c)^{\top} \bar u^c = M_c(f)$,
since $u^c$ has full row rank. Moreover $B\, (\bar u^t \otimes u^{c'}) = \bar u^c$ by
\eqref{eq:min-above}, and $\bar u^t \otimes u^{c'}$ has full row rank, so the rows
of $B$ satisfy exactly the linear relations of the rows of $\bar u^c$. Hence $S$
is the lexicographically first row basis of $\bar u^c$, and $Q^{\top}$ holds the
coordinates of the rows of $\bar u^c$ in the rows indexed by $S$. Choosing
$d(c)$ assignments $\tau_1, \dots, \tau_{d(c)}$ on which the columns of~$u^c$ are independent, the entries of $\bar u^c$ solve a $d(c) \times d(c)$
linear system with entries from $u^c$ and $f$; choosing similarly $|S|$
columns of $\bar u^c$, the entries of $Q$ solve an $|S| \times |S|$ system with
entries from $\bar u^c$. By Cramer's rule, both are ratios of determinants of
matrices with entries of polynomial bit size. So every number written in
Pass~2 has polynomial bit size, and every matrix factored in the procedure
has entries of polynomial bit size, on which Gaussian elimination over $\Q$
runs in polynomial time.
\end{proof}

The size of other circuits structured by vtrees, such as TDDs or d-SDNNF circuits, depends more tightly on the subfunction count $S_t$ than on $\rho_t$. Lemma~\ref{lem:sandwich} bounds the gap between the two parameters.

\begin{lemma}[$\dagger$]\label{lem:sandwich}
For every Boolean function $f \not\equiv 0$ and every vtree node $t$,
$\rho_t(f) \le S_t(f) \le 2^{\rho_t(f)} - 1$.
\end{lemma}
\begin{proof}
Let $\rho = \rho_t(f)$. The rank of $M_t(f)$ is the dimension of the span of its rows, and this span is generated by the $S_t(f)$ distinct nonzero rows, so $\rho \le S_t(f)$.

For the upper bound, choose $\rho$ linearly independent columns of $M_t(f)$, at indices $\sigma_1,\dots,\sigma_\rho$, and map each row $f[\tau]$ to its restriction $(f[\tau](\sigma_1),\dots,f[\tau](\sigma_\rho)) \in \{0,1\}^\rho$. These columns are a basis of the column space, so every column of $M_t(f)$ is a linear combination of them, with coefficients that do not depend on $\tau$. Hence the entries of a row are determined by its restriction: two rows with the same restriction are equal, and a row whose restriction is zero is entirely zero. So the map is injective on the distinct nonzero rows and takes values in $\{0,1\}^\rho \setminus \{0\}$, which gives $S_t(f) \le 2^\rho - 1$.
\end{proof}

The bounds are tight. The upper end is for instance attained by the multiplexer function $\MUX_n$. $\MUX_n$ has $n = 2^k$ data variables and $k$ selector variables and outputs the data variable that the selectors address. For vtrees that have a node $t$ such that $X_t$ is exactly the set of data variables, so a cut separating the data variables from the selectors, we have $S_t = 2^n - 1$ while $\rho_t = n$, as shown in \suppl.

\subsection*{Structured Circuits as TTN}

We recall the definition of TDDs \citep{CapelliEtAl26} and d-SDNNF circuits and explain that these can be seen as TTN. Both TDDs and d-SDNNF are circuits that respect a vtree~$T$.

A \emph{TDD} respecting $T$ places at each vtree node $t$ a set of nodes $N_t = \{v^t_1,v^t_2,\dots\}$, each $v \in N_t$ computing a Boolean function over $X_t$: a leaf node stores its function; an internal node $v \in N_t$  stores a set $E(v)$ of \emph{input pairs} of nodes at the two children and computes $\bigvee_{(v_1, v_2)} f_{v_1} \wedge f_{v_2}$. Determinism requires that distinct nodes at the same vtree node have disjoint models; hence every assignment satisfies at most one input pair. In TDD, determinism is enforced syntactically by ensuring that every pair $(v_1,v_2)$ is an input pair for at most one node. The root carries one node, computing $f$. The width at $t$ is the number of nodes; if $f \not\equiv 0$, the nodes at $t$ of the minimal TDD are the indicators of the classes of assignments inducing the same nonzero subfunction, so their number is $S_t(f)$. TDDs can be seen as TTNs: for every internal node $t$ we have a $|N_t| \times |N_{t_1}| \times |N_{t_2}|$ tensor defined as $W^t_{ijk} = 1$ if $(v^{t_1}_j,v^{t_2}_k) \in E(v^t_i)$ and $0$ otherwise.

\begin{proposition}[$\dagger$]\label{proposition:simulation}
Every TDD of width $k$ respecting $T$ and computing a function $f \not\equiv 0$ can be turned into a TTN with $0/1$ transfer tensors which computes $f$ and has the same width.
\end{proposition}
\begin{proof}
Consider a TDD of width $k$ respecting $T$ and computing $f$, and write $f_v$ for the function computed by the node $v$. Every $N_t$ is non-empty. Otherwise no node at the parent of $t$ has an input pair, so every node there computes $0$; a node whose input pairs all contain a node computing $0$ computes $0$ as well, so the root computes $0$, contradicting $f \not\equiv 0$.

We define the TTN by $d(t) = |N_t|$ for every node $t$, by $u^t_i = f_{v^t_i}$ if $t$ is a leaf, seeing $f_{v^t_i}$ as a $0/1$ row of the leaf matrix, and, if $t$ is internal with children $t_1$ and $t_2$, by $W^t_{ijk} = 1$ if $(v^{t_1}_j,v^{t_2}_k) \in E(v^t_i)$ and $W^t_{ijk} = 0$ otherwise. Then $d(r) = 1$ and $\max_t d(t) = k$, and the transfer tensors are $0/1$.

We claim that $u^t_i = f_{v^t_i}$ for every node $t$ and every $i \in [d(t)]$. For the leaves, this is the definition. Let $t$ be internal with children $t_1$ and $t_2$, and suppose that the claim holds at $t_1$ and $t_2$. By (\ref{eq:evaluate_ttn}), for every $\tau \in 2^{X_t}$ with restrictions $\tau_1$ and $\tau_2$,
\begin{align*}
u^t_i(\tau) = \sum_{(v^{t_1}_j,v^{t_2}_k) \in E(v^t_i)} f_{v^{t_1}_j}(\tau_1)\, f_{v^{t_2}_k}(\tau_2).
\end{align*}
Each term is $1$ if $\tau$ satisfies the input pair $(v^{t_1}_j,v^{t_2}_k)$ and $0$ otherwise. Two distinct pairs cannot both be satisfied by $\tau$: this would make $\tau_1$ a model of both first components, or~$\tau_2$ a model of both second components, with these being distinct nodes at the same vtree node, contradicting determinism. So the sum is $1$ if $\tau$ satisfies some input pair of $v^t_i$ and $0$ otherwise, that is, $u^t_i(\tau) = \bigvee_{(v_1,v_2) \in E(v^t_i)} f_{v_1}(\tau_1) \wedge f_{v_2}(\tau_2) = f_{v^t_i}(\tau)$. This proves the claim. At the root, whose single node computes $f$, it gives $u^r_1 = f$, so the TTN computes $f$.
\end{proof}

Figure~\ref{fig:tdd} gives, on the left, a TDD for $\SUMBIT_2$ and, on the right, its TTN translation. At each node of the tree we show the parent unfolding $\bar W$ rather than $W$: the rows are indexed by $i$, the columns are indexed by $jk$ and the cells contain the value $W_{ijk}$.

Every TDD translates into a circuit in deterministic structured-decomposable negation normal form (d-SDNNF)~\citep{PipatsrisawatDarwiche08}; see also \citet{BovaCapelliMengelSlivovsky15} on compiling CNFs into d-SDNNFs. A \emph{structured DNNF} (SDNNF) respecting $T$ is a circuit of $\lor$-gates and binary $\land$-gates following the vtree: an $\land$-gate at $t$ conjoins circuits over $X_{t_1}$ and $X_{t_2}$; it is deterministic (d-SDNNF) if the inputs of every $\lor$-gate have disjoint sets of models. Hence every $\lor$-gate computes the sum of its inputs and every $\land$-gate their product, so evaluating the circuit with $+$ and $\times$ over $\Q$ at an assignment gives the Boolean value of the circuit at that assignment.

\begin{proposition}\label{proposition:dsdnnf}
Every structured deterministic DNNF (d-SDNNF) made of $s$ gates, respecting $T$, and
computing $f$ gives $\rho_t(f) \le s$ at every node $t$ of $T$.
\end{proposition}
\begin{proof}
We see every gate $g$ of the circuit as a function $f_g$ on $2^X$. Fix a node $t$ of $T$ and let $G_t$ be the set of gates whose variables all lie in $X_t$. For $h \in G_t$, the value of $f_h$ depends only on the assignment to $X_t$, so we read $f_h$ as a vector of $\Q^{2^{X_t}}$. Let $\mathbf{1}$ be the all-ones vector of $\Q^{2^{X_t}}$ and let $U_t = \text{span}_{\Q}\left(\{\mathbf{1}\} \cup \{f_h : h \in G_t\}\right)$, of dimension at most $|G_t| + 1$. We claim that, for every gate $g$ and every $\sigma \in 2^{X \setminus X_t}$, the vector $\tau \mapsto f_g(\tau \times \sigma)$ lies in $U_t$. The claim is shown by induction on the circuit, from its inputs up.

If the variables of $g$ all lie in $X_t$, then the vector is $f_g$ with $g \in G_t$. If $g$ has no variable in $X_t$, then the vector is constant, hence a multiple of $\mathbf{1}$. Otherwise, $g$ is an internal gate. If $g$ is an $\vee$-gate with inputs $g_1,\dots,g_m$, then these inputs have disjoint sets of models by determinism, so $f_g = \sum_{i} f_{g_i}$ and the claim for $g$ follows from the claim for the inputs. If $g$ is an $\wedge$-gate at a node $a$, then its variables all lie in $X_a$, and, since the two previous cases do not apply, at least one of them is in $X_t$ and at least one is not. So $X_a$ intersects $X_t$ without being contained in it. Since the leaf sets of two nodes of $T$ are either disjoint or nested, $X_t \subsetneq X_a$, that is, $a$ is a strict ancestor of $t$. Let $g_1$ be the input of $g$ on the side of $a$ containing $t$ and $g_2$ the other input. The variables of $g_2$ are not in $X_t$, so $f_{g_2}(\tau \times \sigma)$ does not depend on $\tau$, and $f_g(\tau \times \sigma) = f_{g_1}(\tau \times \sigma) f_{g_2}(\tau \times \sigma)$ is a scalar multiple of the vector for~$g_1$ as a function of $\tau$. The claim for $g$ follows from the claim for $g_1$.

Applying the claim to the output gate gives $f[\sigma] \in U_t$ for every $\sigma \in 2^{X \setminus X_t}$, so $V_t(f) \subseteq U_t$ and $\rho_t(f) \le |G_t| + 1$. If the output gate is not in $G_t$, then $|G_t| \le s - 1$ and $\rho_t(f) \le s$. If the output gate is in $G_t$, then $f$ depends only on the variables in $X_t$, so all subfunctions $f[\sigma]$ are equal and $\rho_t(f) \le 1 \le s$.
\end{proof}

Consequently, TTNs polynomially simulate TDDs, d-SDNNFs and all their sublanguages like SDDs~\citep{Darwiche11}. The converse is not true, and not just for some vtree but uniformly for all vtrees. Let $\HWB_n$ be the hidden weighted bit function over $n$ variables $x_1,\dots,x_n$ which outputs for each assignment the variable indexed by its Hamming weight, i.e., $\HWB_n(\tau) = \tau(x_w)$ for $w = \sum_{i} \tau(x_i)$, and $0$ for $w = 0$.

\begin{theorem}[$\dagger$]\label{thm:separation}
For every vtree $T$, $\HWB_n$ has factor rank at most $n + 2$, hence for every vtree~$\HWB_n$ has a TTN representation of width at most $n+2$ and size $O(n^4)$.
\end{theorem}
\begin{proof}
Let $f = \HWB_n$, let $t$ be a node of $T$ and let $Z = X \setminus X_t$. For $\tau \in 2^{X_t}$ and $\sigma \in 2^{Z}$, the Hamming weight of $\tau \times \sigma$ is $w = \wt{\tau} + \wt{\sigma}$, where $\wt{\cdot}$ denotes the Hamming weight of an assignment. If $w \ge 1$, the variable $x_w$ belongs either to $X_t$ or to $Z$, and $f(\tau \times \sigma)$ is read from $\tau$ in the first case and from $\sigma$ in the second. We thus write $M_t(f) = A + B$, where
\begin{align*}
A(\tau,\sigma) = \begin{cases} \tau(x_w) & \text{if } w \ge 1 \text{ and } x_w \in X_t, \\ 0 & \text{otherwise,} \end{cases}
\qquad
B(\tau,\sigma) = \begin{cases} \sigma(x_w) & \text{if } w \ge 1 \text{ and } x_w \in Z, \\ 0 & \text{otherwise.} \end{cases}
\end{align*}
For a fixed $\tau$, the entry $A(\tau,\sigma)$ depends on $\sigma$ only through $\wt{\sigma} \in \{0,\dots,|Z|\}$. So two columns of $A$ indexed by assignments of the same Hamming weight are equal, $A$ has at most $|Z| + 1$ distinct columns, and $\rk A \le |Z| + 1$. Symmetrically, $B(\tau,\sigma)$ depends on $\tau$ only through $\wt{\tau} \in \{0,\dots,|X_t|\}$, so $B$ has at most $|X_t| + 1$ distinct rows and $\rk B \le |X_t| + 1$. Hence
\begin{align*}
\rho_t(f) = \rk M_t(f) \le \rk A + \rk B \le |Z| + |X_t| + 2 = n + 2.
\end{align*}
This holds at every node $t$, so $\fr(f,T) \le n + 2$. By Proposition~\ref{prop:realization}, there is a TTN computing~$f$ with $d(t) = \rho_t(f) \le n + 2$ at every node, of size $O(|X| \cdot \fr(f,T)^3) = O(n^4)$.
\end{proof}
Since there is a constant $c > 0$ such that, for every multiple $n$ of $7$, every TDD for $\HWB_n$, over every vtree, has size at least $2^{cn}$ \citep[Theorem~21]{CapelliEtAl26}, TTNs are exponentially more succinct than TDDs. \citet{OnakaEtAl25tt} give a rank-$2n$ tensor train for $\HWB_n$ on a linear order, or equivalently, linear vtree; our bound holds for every vtree.

Unlike d-SDNNF circuits, TTNs support polynomial-time negation (see the proof of Theorem~\ref{thm:operations}), and, since $M_t(\neg f) = J - M_t(f)$ for the all-ones matrix $J$, negation increases each cut rank by at most one. With Theorem~2 of \citet{VinallSmeeth24}, this gives a quasi-polynomial separation of TTNs from d-SDNNF circuits. For every $s$ there is a Boolean function $f$ with a d-SDNNF of size $s$ such that every SDNNF computing $\neg f$, on any vtree, has size $s^{\tilde\Omega(\log s)}$, whereas, by Proposition~\ref{proposition:dsdnnf},~$\neg f$ has a TTN of width at most $s + 1$ on the vtree of that d-SDNNF.

\section{A Canonical Form for TTN}\label{sec:canon}

\subsection{Definition}

In this section, we define our main object, a canonical TTN that we call ATDD. Fix a node $t$ in the vtree and let $\rho$ be the dimension of the cut space $V_t(f)$. We consider the subfunctions $f[\tau]$ obtained for every assignment $\tau$ to $X_t$. Imagine reading these assignments in the lexicographic order induced by the vtree's leaves. We do not keep all assignments during the enumeration: assignment $\tau \in 2^{X_t}$ is kept if and only if the subfunction $f[\tau]$ is non-zero and is not a $\mathbb{Q}$-linear combination of the subfunctions already kept. We call $\tau$ a \emph{canonical test}. We thus obtain a list of $\rho$ assignments in lexicographic order 
$$p^t_1 < \dots < p^t_{\rho}.$$
Thus, for each $\tau \in 2^{X_t}$ we have a unique sequence $(e^t_1(\tau),\dots,e^t_{\rho}(\tau))$ such that
\begin{equation}\label{eq:linear_combination}
f[\tau] = \sum_{i = 1}^{\rho} f[p^t_i]e^t_i(\tau).
\end{equation}
Each $e^t_i$ is seen as a vector of length $2^{|X_t|}$, or a function in $2^{X_t} \rightarrow \mathbb{Q}$. Since $(f[p^t_1],\dots,f[p^t_\rho])$ is a basis of $\text{span}_{\mathbb{Q}}(f[\tau] : \tau \in 2^{X_t})$, $(e^t_1,\dots,e^t_\rho)$ is its \emph{dual basis} for $V_t(f) = \text{span}_{\mathbb{Q}}(f[\sigma] : \sigma \in 2^{X\setminus X_t})$ and we have $v(\tau) = \sum_i v(p_i)e^t_i(\tau)$ for every $v \in V_t(f)$ and $\tau \in 2^{X_t}$. We call $e^t_1,\dots,e^t_\rho$ the \emph{interpolators} at $t$.

\begin{example}\label{ex:sumbitcanon}
We continue Example~\ref{ex:sumbit} at the node $t$. Scanning the subfunctions for the
assignments $\bar x_1\bar x_2, \bar x_1 x_2,  x_1\bar x_2, x_1 x_2$, we keep the first three
and discard $f[x_1 x_2] = f[\bar x_1\bar x_2] - f[\bar x_1 x_2] + f[x_1\bar x_2]$, so the canonical tests are $p^t_1 =
\bar x_1\bar x_2$, $p^t_2 = \bar x_1 x_2$, $p^t_3 = x_1\bar x_2$ and $\rho_{t} = 3$. We then have
$e^{t}_1 = (1, 0, 0, 1)$,
$e^{t}_2 = (0, 1, 0, -1)$,
$e^{t}_3 = (0, 0, 1, 1)$.
\end{example}

\begin{example}
Say $\ell$ is the leaf labeled with the variable $x$, and assume $f \neq 0$. The two subfunctions~$f[\bar x]$ and $f[x]$ are $0/1$ vectors, so they are linearly dependent only if they are equal or if one of them is zero. If $f[\bar x] = f[x]$, the single canonical test is $p^\ell_1 = \bar x$ and $e^\ell_1 = (1, 1)$; if $f[x] = 0$, then $p^\ell_1 = \bar x$ and $e^\ell_1 = (1, 0)$; if $f[\bar x] = 0$, then $p^\ell_1 = x$ and $e^\ell_1 = (0, 1)$. If $\rho = 2$, the canonical tests are $p^\ell_1 = \bar x$,~$p^\ell_2 = x$, and $e^\ell_1 = (1, 0)$, $e^\ell_2 = (0, 1)$.
\end{example}

When $t = r$ is the root of the vtree, $\rho$ is $1$ and $f[p^r_1] = f(p^r_1) = 1$ (assuming $f$ is non-zero) thus (\ref{eq:linear_combination}) becomes
\begin{equation}\label{eq:f_is_e}
f = e^r_1.
\end{equation}
Our objective is to represent vectors $e^t_1,\dots,e^t_{\rho}$ for all $t$. Each vector for $t$ is potentially too long to store but can be succinctly represented given the vectors for the children of $t$. Suppose $t$ has children~$t_1$ and $t_2$, and that we know the children canonical tests $p^{t_1}_1,\dots,p^{t_1}_{\rho_1}$ and $p^{t_2}_1,\dots,p^{t_2}_{\rho_2}$, with~$\rho_1$ and $\rho_2$ the dimensions of $V_{t_1}$ and $V_{t_2}$. We claim that storing the entries of $e^t_i$ at each combination $p^{t_1}_j \times p^{t_2}_k$ is sufficient. To see that, we first recall that $V_t$ is a subspace of the tensor product of $V_{t_1}$ and $V_{t_2}$, denoted by $V_{t_1} \otimes V_{t_2}$ (Proposition~\ref{proposition:nestedness}). Consider the basis $(e^{t_1}_j \otimes e^{t_2}_k)_{j,k}$ of $V_{t_1} \otimes V_{t_2}$ defined by $(e^{t_1}_j \otimes e^{t_2}_k)(\tau_1 \times \tau_2) = e^{t_1}_j(\tau_1)e^{t_2}_k(\tau_2)$. By Proposition~\ref{proposition:nestedness}, there are $\rho_1\rho_2$ unique coefficients $\alpha^t_{i,j,k}$ such that
\begin{equation}\label{eq:base_combination}
e^t_i = \sum_{j,k} \alpha^t_{i,j,k} e^{t_1}_j \otimes e^{t_2}_k.
\end{equation}
Evaluating both sides on $p^{t_1}_j \times p^{t_2}_k$ and observing that $(e^{t_1}_{j'} \otimes e^{t_2}_{k'})(p^{t_1}_j \times p^{t_2}_k) = 1$ if $j = j'$ and $k = k'$, and $0$ otherwise, we get that $e^t_i(p^{t_1}_j \times p^{t_2}_k) = \alpha^t_{i,j,k}$. ATDDs thus store the $\rho_1 \times \rho_2$ coefficients $\alpha^t_{i,j,k}$ in the transfer tensors so that, if we can recursively evaluate $e^{t_1}_j$ and $e^{t_2}_k$, then we get $e^t_i$ through (\ref{eq:base_combination}).

\begin{definition}[ATDD]\label{def:canonical}
The \emph{ATDD} of the $0$ function with respect to $T$ is a single node labeled $0$. 
The \emph{ATDD} of a Boolean function $f \not\equiv 0$ with respect to $T$
is a tree tensor network with $d(t) = \rho_t(f)$ for every node $t$ of $T$. 
If $t$ is a leaf, then its bond vectors are $u^t_i = e^t_i$ for every $i \in [d(t)]$. 
If $t$ is an internal node with children $t_1$ and $t_2$ then its transfer tensor is defined by $W^t_{ijk} = e^t_i(p^{t_1}_j \times p^{t_2}_k)$ for every
$(i,j,k) \in [d(t)] \times [d(t_1)] \times [d(t_2)]$.
\end{definition}

\begin{restatable}[$\dagger$]{theorem}{soundness}\label{thm:sound}
The ATDD of a Boolean function $f \not\equiv 0$ is well defined and computes $f$.
\end{restatable}
\begin{proof}
For every node $t$ in the vtree, every $i \in [d(t)]$ and every assignment $\tau$ to $X_t$, we claim that $u^t_i(\tau) = e^t_i(\tau)$; with (\ref{eq:f_is_e}) it implies that the ATDD computes $f$. The claim is shown by induction. The base case holds by the definition of the ATDD bond vectors at the leaves (Definition~\ref{def:canonical}). For the inductive case, suppose $t$ has two children $t_1$ and $t_2$. By (\ref{eq:evaluate_ttn})
\begin{align*}
u^t_i(\tau) = \sum_{j = 1}^{d(t_1)} \sum_{k = 1}^{d(t_2)} W^t_{ijk} u^{t_1}_j(\tau_1) u^{t_2}_k(\tau_2).
\end{align*}
By induction, 
\begin{align*}
u^{t_1}_j(\tau_1) u^{t_2}_k(\tau_2) = e^{t_1}_j(\tau_1) e^{t_2}_k(\tau_2) = (e^{t_1}_j \otimes e^{t_2}_k)(\tau).
\end{align*} 
Using (\ref{eq:base_combination}) and that $W^t_{ijk} = e^t_i(p^{t_1}_j \times p^{t_2}_k) = \alpha^t_{i,j,k}$, it follows that $u^t_i(\tau) = e^t_i(\tau)$, as claimed.
\end{proof}

For a nonzero function $f$ from $2^X$ to $\Q$ that is not Boolean, the same construction gives $f = f(p^r_1)\, e^r_1$ in place of \eqref{eq:f_is_e}, so the network of Definition~\ref{def:canonical}, built from the same tests and interpolators, computes $f/f(p^r_1)$, and the scalar $f(p^r_1)$ would be kept separately.

The tensors and vectors defining an ATDD are totally and deterministically defined by the pair~$(f,T)$. Thus there is a unique ATDD respecting $T$ that represents $f$. Hence, ATDDs provide canonical representations for Boolean functions (for a fixed vtree). 

\begin{theorem}[$\dagger$]\label{thm:canonical}
Two ATDDs respecting the same vtree
$T$ and computing non-zero Boolean functions 
compute the same function if and only if they are
syntactically identical.
\end{theorem}

We remark that $T$ serves to define the lexicographic ordering on the assignments and that using different orderings would change the tensors or vectors of the ATDD but not their dimensions, which only depend on the dimension of the cut spaces $V_t(f)$.

\begin{figure}[t]
\centering
\includegraphics[width=\wfmtA\columnwidth]{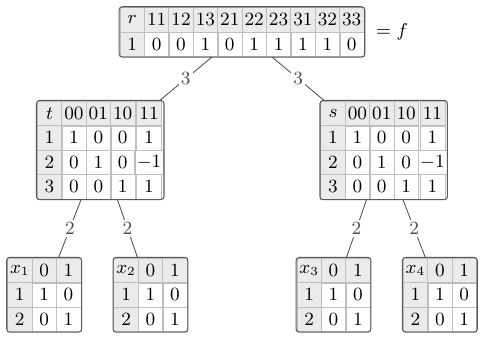}
\vspace{0pt}
\caption{The ATDD of $\SUMBIT_2$ over the same vtree used in
Figure~\ref{fig:tdd}. The ATDD is drawn as the vtree annotated with
tensor tables. Each node is one table: a leaf $x_i$ holds its $2\times
2$ matrix (columns are the values $0,1$); an internal node holds
the parent unfolding $\bar W^t$ of its transfer tensor, with rows $1,\dots,d(t)$ and
columns the products of the children's canonical tests in lexicographic order, that is, the
assignments $00,01,10,11$ to the two child variables at $t,s$, and
the pairs $jk$ of the children's canonical tests at the root, whose
single row records the values of $f$ at these products of tests. Each edge is labeled by the child bond
dimension, so the whole ATDD (vtree, bond dimensions, leaf matrices,
transfer tensors) is recoverable from the picture. Row $2$ of the tables
at $t$ and $s$ carries a $-1$ entry (Example~\ref{ex:sumbitcanon}).}
\label{fig:atdd}
\end{figure}

\subsection{Identifying ATDD}

There can be several tree tensor networks respecting $T$ and representing $f$. We here explain that ATDD form can be checked locally. We claim that, for every vtree node $t$, one only has to check that the tensor $W^t$ has a specific shape. As explained before, the entries of $W^t$ are the entries of the vectors $e^t_i$ at compositions of the children canonical tests. It turns out the canonical tests of $t$ can be expressed as composition of the children canonical tests.

\begin{lemma}[$\dagger$]\label{lem:compose}
The canonical tests for an internal node $t$ are compositions of the canonical tests of its children $t_1$ and $t_2$. For each $i \in [\rho_t]$, there is $(j,k) \in [\rho_{t_1}] \times [\rho_{t_2}]$ such that $p^t_i = p^{t_1}_j \times p^{t_2}_k$.
\end{lemma}
\begin{proof}
We first observe that, at any node, an assignment $\tau$ is not a canonical test as soon as $f[\tau]$ is a linear combination of subfunctions $f[\tau']$ with $\tau' < \tau$. Indeed, when the scan reaches $\tau$, each such~$f[\tau']$ is either a kept subfunction, or zero, or a linear combination of kept subfunctions. So $f[\tau]$ is a linear combination of the subfunctions already kept (possibly the empty one, if $f[\tau] = 0$) and $\tau$ is discarded. In the same way, if $\tau$ is not a canonical test, then $f[\tau]$ is a linear combination of canonical tests' subfunctions $f[p]$ with $p < \tau$.

Let $\tau = \tau_1 \times \tau_2$ with $\tau_1 \in 2^{X_{t_1}}$ and $\tau_2 \in 2^{X_{t_2}}$. Since the leaves of $X_{t_1}$ precede those of~$X_{t_2}$, the order on $2^{X_t}$ is the lexicographic order on the pairs $(\tau_1,\tau_2)$. We show that if $\tau_1$ is not a canonical test of $t_1$ or $\tau_2$ is not a canonical test of $t_2$, then $\tau$ is not a canonical test of~$t$. Suppose first that $\tau_1$ is not a canonical test of $t_1$, so $f[\tau_1] = \sum_{j : p^{t_1}_j < \tau_1} \lambda_j f[p^{t_1}_j]$ for some coefficients $\lambda_j$. These vectors are indexed by $2^{X \setminus X_{t_1}}$. Evaluating them at $\tau_2 \times \sigma$ for every $\sigma \in 2^{X \setminus X_t}$ gives
\begin{align*}
f[\tau_1 \times \tau_2] = \sum_{j : p^{t_1}_j < \tau_1} \lambda_j f[p^{t_1}_j \times \tau_2].
\end{align*}
Since $p^{t_1}_j < \tau_1$, we have $p^{t_1}_j \times \tau_2 < \tau_1 \times \tau_2$, so $\tau$ is not a canonical test of $t$ by the first observation. Suppose now that $\tau_2$ is not a canonical test of $t_2$, so $f[\tau_2] = \sum_{k : p^{t_2}_k < \tau_2} \lambda_k f[p^{t_2}_k]$. Evaluating at $\tau_1 \times \sigma$ gives
\begin{align*}
f[\tau_1 \times \tau_2] = \sum_{k : p^{t_2}_k < \tau_2} \lambda_k f[\tau_1 \times p^{t_2}_k].
\end{align*}
Since $p^{t_2}_k < \tau_2$, we have $\tau_1 \times p^{t_2}_k < \tau_1 \times \tau_2$, so again $\tau$ is not a canonical test of $t$. Hence, every canonical test of $t$ is of the form $p^{t_1}_j \times p^{t_2}_k$.
\end{proof}

Since $e^t_i$ is $1$ on $p^t_i$ and $0$ on all other canonical tests, the entry $W^t_{ijk}$ is $1$ if $p^t_i = p^{t_1}_j \times p^{t_2}_k$ but it is~$0$ if $p^t_{i'} = p^{t_1}_j \times p^{t_2}_k$ for some $i' \neq i$. A discarded subfunction $f[\tau]$ is a combination of the tests kept before $\tau$ only, so $e^t_i(\tau) = 0$ for every $\tau < p^t_i$; hence the first nonzero entry of row $i$ of $\bar W^t$ is the $1$ at the column of $p^t_i$, since the columns $(j,k)$ of $\bar W^t$ are listed in the order of the assignments $p^{t_1}_j \times p^{t_2}_k$ (the order on $2^{X_t}$ is lexicographic on pairs, as in the proof of Lemma~\ref{lem:compose}). These pivot columns increase with $i$, because $p^t_1 < \dots < p^t_{\rho_t}$. So the $\rho_t \times \rho_{t_1}\rho_{t_2}$ parent unfolding $\bar W^t$ is in reduced row echelon form.

\begin{example}
Continuing Example~\ref{ex:sumbit} at the node $t$, we have $(p^{t_1}_1,p^{t_1}_2) = (\colorbox{red!40}{$0$},\colorbox{red!40}{$1$})$, $(p^{t_2}_1,p^{t_2}_2) = (\colorbox{blue!40}{$0$},\colorbox{blue!40}{$1$})$ and $(p^t_1,p^t_2,p^t_3) = (00,01,10)$, thus
{\scriptsize\fboxsep2pt
\begin{align*}
W^t &= 
\left( 
\begin{pmatrix} 
e^t_1(\colorbox{red!40}{$0$}\colorbox{blue!40}{$0$}) & e^t_1(\colorbox{red!40}{$0$}\colorbox{blue!40}{$1$}) 
\\ 
e^t_1(\colorbox{red!40}{$1$}\colorbox{blue!40}{$0$}) & e^t_1(\colorbox{red!40}{$1$}\colorbox{blue!40}{$1$}) 
\end{pmatrix}, 
\begin{pmatrix}  
e^t_2(\colorbox{red!40}{$0$}\colorbox{blue!40}{$0$}) & e^t_2(\colorbox{red!40}{$0$}\colorbox{blue!40}{$1$}) 
\\
e^t_2(\colorbox{red!40}{$1$}\colorbox{blue!40}{$0$}) & e^t_2(\colorbox{red!40}{$1$}\colorbox{blue!40}{$1$}) 
\end{pmatrix},
\begin{pmatrix} 
e^t_3(\colorbox{red!40}{$0$}\colorbox{blue!40}{$0$}) & e^t_3(\colorbox{red!40}{$0$}\colorbox{blue!40}{$1$}) 
\\
e^t_3(\colorbox{red!40}{$1$}\colorbox{blue!40}{$0$}) & e^t_3(\colorbox{red!40}{$1$}\colorbox{blue!40}{$1$}) 
\end{pmatrix}
\right)
\\
\bar W^t &=  
\begin{pmatrix} 
e^t_1(\colorbox{red!40}{$0$}\colorbox{blue!40}{$0$}) & e^t_1(\colorbox{red!40}{$0$}\colorbox{blue!40}{$1$}) & e^t_1(\colorbox{red!40}{$1$}\colorbox{blue!40}{$0$}) & e^t_1(\colorbox{red!40}{$1$}\colorbox{blue!40}{$1$}) 
\\
e^t_2(\colorbox{red!40}{$0$}\colorbox{blue!40}{$0$}) & e^t_2(\colorbox{red!40}{$0$}\colorbox{blue!40}{$1$}) & e^t_2(\colorbox{red!40}{$1$}\colorbox{blue!40}{$0$}) & e^t_2(\colorbox{red!40}{$1$}\colorbox{blue!40}{$1$}) 
\\
e^t_3(\colorbox{red!40}{$0$}\colorbox{blue!40}{$0$}) & e^t_3(\colorbox{red!40}{$0$}\colorbox{blue!40}{$1$}) & e^t_3(\colorbox{red!40}{$1$}\colorbox{blue!40}{$0$}) & e^t_3(\colorbox{red!40}{$1$}\colorbox{blue!40}{$1$}) 
\end{pmatrix}=  
\begin{pmatrix} 
1 & 0 & 0 & 1
\\
0 & 1 & 0 & -1 
\\
0 & 0 & 1 & 1
\end{pmatrix}
\end{align*}
}
\end{example}

For a leaf $\ell$, the leaf matrix $u^\ell$ is already in reduced row echelon form: by the leaf example above it is one of ${\scriptsize\fboxsep2pt\begin{pmatrix} 1 & 0 \\ 0 & 1 \end{pmatrix}}$, $(1\ 1)$, $(1\ 0)$, $(0\ 1)$. We thus define $\bar W^\ell = u^\ell$. 

An ATDD is thus locally characterized by the shape of its tensors, and the characterization can be checked on the stored numbers alone. Besides the parent unfolding $\bar W^t$, the characterization uses the child unfoldings $W^t_{\langle t_1 \rangle}$ and $W^t_{\langle t_2 \rangle}$ introduced for the minimization procedure (Section~\ref{sec:width}).

\begin{theorem}[local characterization, $\dagger$]\label{thm:local}
Let $A$ be a TTN respecting $T$ that computes a Boolean function~$f \not\equiv 0$. Then $A$ is the ATDD of $f$ if and only if, at every node $t$ of $T$,
\begin{enumerate}
\item[(i)] $\bar W^t$ is in reduced row echelon form and has no zero row, and
\item[(ii)] if $t$ is internal, both child unfoldings $W^t_{\langle t_1 \rangle}$ and $W^t_{\langle t_2 \rangle}$ have full row rank.
\end{enumerate}
\end{theorem}

Condition (i) makes the bond vectors at $t$ linearly independent and fixes their basis, bottom-up; condition (ii) makes the context vectors of $t_1$ and $t_2$ linearly independent, top-down. Together they force $d(t) = \rho_t(f)$ at every node, so minimality is a consequence of the two conditions rather than a hypothesis.

\begin{proof}
We use the context vectors, the rank criterion and the two facts about full row rank of Section~\ref{sec:width}. By the rank criterion, $A$ is minimal if and only if $u^t$ and $\bar u^t$ have full row rank at every node $t$. We also use the statement for minimal TTNs proved in \suppl, namely that a minimal TTN computing a Boolean function $f \not\equiv 0$ is the ATDD of $f$ if and only if every $\bar W^t$ is in reduced row echelon form.

\emph{If.} Assume (i) and (ii) at every node. We first show bottom-up that every $u^t$ has full row rank. At a leaf, $u^t = \bar W^t$ is in reduced row echelon form without zero rows, so its rows are independent. At an internal node $t$, $u^t = \bar W^t (u^{t_1} \otimes u^{t_2})$ by \eqref{eq:evaluate_ttn}. Here $\bar W^t$ has full row rank by (i), and $u^{t_1} \otimes u^{t_2}$ by induction and the first fact, so $u^t$ has full row rank by the second. We then show top-down that every $\bar u^t$ has full row rank. At the root, $\bar u^r = (1)$ has full row rank. At an internal node $t$, $\bar u^{t_1} = W^t_{\langle t_1 \rangle} (\bar u^t \otimes u^{t_2})$ by \eqref{eq:min-above}. Here $W^t_{\langle t_1 \rangle}$ has full row rank by~(ii), and $\bar u^t \otimes u^{t_2}$ by the first fact, since $\bar u^t$ (by induction) and~$u^{t_2}$ (by the first step) have full row rank and live on the disjoint sets $X \setminus X_t$ and $X_{t_2}$. So $\bar u^{t_1}$ has full row rank, and symmetrically $\bar u^{t_2}$. By the rank criterion, $A$ is minimal, and by (i) and the statement for minimal TTNs it is the ATDD of $f$.

\emph{Only if.} Let $A$ be the ATDD of $f$. Every $\bar W^t$ is in reduced row echelon form, as shown before the theorem. The ATDD is minimal by Definition~\ref{def:canonical}, so by the rank criterion $u^t$ and $\bar u^t$ have full row rank at every node $t$. A zero row of $\bar W^t$ would give a zero row of $u^t$, directly at a leaf and by \eqref{eq:evaluate_ttn} at an internal node, so (i) holds. If $x^{\top} W^t_{\langle t_1 \rangle} = 0$ for some $x \ne 0$, then $x^{\top} \bar u^{t_1} = 0$ by \eqref{eq:min-above}, which contradicts the full row rank of $\bar u^{t_1}$. So $W^t_{\langle t_1 \rangle}$ has full row rank, symmetrically $W^t_{\langle t_2 \rangle}$, and (ii) holds.
\end{proof}

Condition (ii) cannot be dropped. For example, the TTN of Figure~\ref{fig:tdd} has every $\bar W^t$ in reduced row echelon form without zero rows, but at the root its child unfoldings are the $4 \times 4$ cut matrix $M_t(f)$ and its transpose, of rank $3 < 4$, so not of full row rank, and indeed the TTN is not minimal. The verification costs one check of reduced row echelon form per node and two rank computations per internal node. Verifying that a matrix is in reduced row echelon form is linear in its size, and the rank of a $w \times w^2$ matrix is computed by Gaussian elimination in $O(w^4)$ arithmetic operations, so the conditions of Theorem~\ref{thm:local} are checked with $O(|X| \cdot w^4)$ arithmetic operations for a TTN of width $w$, without reference to $f$. Every TTN computing a function $f \not\equiv 0$ can be made minimal (Proposition~\ref{prop:minimization}), and a minimal TTN computing a Boolean function can then be brought to ATDD form in polynomial time.

\begin{theorem}\label{thm:makeCanonical}
There is an algorithm that, given a minimal TTN respecting $T$ and computing a Boolean function~$f \not\equiv 0$ returns the ATDD for $f$ respecting $T$ in polynomial time.
\end{theorem}
\begin{proof}
The algorithm changes the basis of the bond vectors at one node at a time and compensates in the parent, so that the computed function never changes. Let $t$ be a non-root node with parent $p$, say the left child, and let $\Gamma \in \Q^{d(t) \times d(t)}$ be invertible. Replace $\bar W^t$ by $\Gamma\, \bar W^t$ (at a leaf, $u^t$ by $\Gamma\, u^t$), which by \eqref{eq:evaluate_ttn} replaces the rows of $u^t$, the bond vectors at $t$, by $u'^t_a = \sum_b \Gamma_{ab}\, u^t_b$, and replace $W^p$ by $W'^p_{iak} = \sum_{j} W^p_{ijk}\, (\Gamma^{-1})_{ja}$. Then $\sum_a W'^p_{iak}\, u'^t_a = \sum_j W^p_{ijk}\, u^t_j$, so $u^p$ is unchanged, and so is every bond vector above $p$ and $f$ itself. We call this a \emph{local move} at $t$; it changes only the tensors at $t$ and $p$ and the bond vectors at $t$.

Recall that the interpolators at $t$ are the dual basis vectors $e^t_1, \dots, e^t_{\rho_t}$ of Section~\ref{sec:canon}. Since the TTN is minimal, the rows of $u^t$ are a basis of $V_t$ at every node $t$ (by \eqref{eq:min-split} and the rank criterion of Section~\ref{sec:width}), and a local move keeps this. Process the non-root nodes in post-order, maintaining the invariant that the bond vectors of every processed node are the interpolators; later local moves do not change it, since they touch only the node processed and its parent. At a leaf $t$, take $\Gamma$ with $\Gamma\, u^t$ in reduced row echelon form, invertible since the rows of $u^t$ are independent, and apply the local move. At an internal~$t$ both children are processed, so by~\eqref{eq:evaluate_ttn} the rows of $\bar W^t$ are the coefficients of the bond vectors $u^t_i$ in the product family $e^{t_1}_j \otimes e^{t_2}_k$, listed in lexicographic $(j,k)$ order; the rows of $\bar W^t$ are independent because the $u^t_i$ are, so again $\Gamma$ with $\Gamma\, \bar W^t$ in reduced row echelon form is invertible; apply the local move. At the root nothing is done, since $d(r) = 1$, and the single row of $\bar W^r$ is the coefficient row of~$f$, whose first nonzero entry is the value of $f$ at its lexicographically first model, which is $1$ because $f$ is Boolean (by the property shown in the next paragraph).

The invariant holds after the step at $t$. Since the leaves of $X_{t_1}$ precede those of $X_{t_2}$, the order on $2^{X_t}$ is the lexicographic order on pairs $\tau_1 \times \tau_2$, so $e^{t_1}_j \otimes e^{t_2}_k$ vanishes at every assignment before $p^{t_1}_j \times p^{t_2}_k$ and takes the value $1$ there; and by the computation after \eqref{eq:base_combination} it takes the value $0$ at every other product of tests. Hence a vector $v = \sum_{j,k} \alpha_{jk}\, e^{t_1}_j \otimes e^{t_2}_k$ vanishes before the first nonzero product of tests in its coefficient row and equals that coefficient there. The rows of $\bar W^t$ are the coefficient rows of a basis of $V_t \subseteq V_{t_1} \otimes V_{t_2}$, so the rows of $\Gamma\, \bar W^t$ are the coefficient rows of vectors of $V_t$ that vanish before the product of tests at their pivot column, their \emph{pivot product}, take the value $1$ at it, and $0$ at the pivot products of the other rows. These vectors are the interpolators $e^t_i$ and the pivot products are the tests $p^t_i$, written as pairs $(j,k)$, since a vector of $V_t$ vanishing before a test $p^t_i$ and taking prescribed values at the tests is determined by \eqref{eq:linear_combination}, and the pivots of a basis of $V_t$ in reduced row echelon form, the lexicographically first assignments at which its vectors are nonzero, are the canonical tests of $t$ (\suppl). After the pass, every bond vector is an interpolator, so by the uniqueness of the coefficients in \eqref{eq:base_combination} every tensor is $W^t_{ijk} = e^t_i(p^{t_1}_j \times p^{t_2}_k)$, and $u^r_1 = f = e^r_1$, so the TTN is the ATDD of Definition~\ref{def:canonical}.

Each step computes a reduced row echelon form of a matrix with at most $w$ rows and at most $\max(w^2, 2)$ columns, reads off $\Gamma^{-1}$ as the restriction of the matrix to its pivot columns, and folds a~$w \times w$ matrix into a tensor with at most $w^3$ entries, where $w$ is the width, which the pass does not change; both take $O(w^4)$ arithmetic operations, so the algorithm runs with $O(|X| \cdot w^4)$ arithmetic operations. The numbers stay small: the steps below $t$ do not change~$u^t$, and after the step at $t$ the bond vectors are the interpolators, so $\Gamma_t\, u^t = e^t$; evaluating at the canonical tests gives $\Gamma_t H_t = I$ with $H_t = (u^t_i(p^t_j))_{ij}$, so $\Gamma_t = H_t^{-1}$ where $H_t$ holds values of the input's bond vectors, computable by contraction with polynomial bit size, and every tensor written is an input tensor multiplied by such matrices and their inverses.
\end{proof}

In concurrent work, \citet{Vilmart26} gives a strict canonical form for tensor trains (the case of a linear vtree) over arbitrary fields, by echelon forms. ATDDs give a strict canonical form over arbitrary vtrees. 

\subsection*{Seeing TDDs as ATDDs}

ATDDs are backward compatible with canonical TDDs. Call $t$ a \emph{counting node} if
$\rho_t = S_t$. The \emph{subfunction classes} at $t$ group the assignments $\tau \in 2^{X_t}$ with $f[\tau] \neq 0$ by the value of $f[\tau]$.

\begin{proposition}[$\dagger$]\label{prop:counting}
At a counting node the canonical tests are the lexicographically least
members of the subfunction classes and the interpolators are their
$0/1$ indicator vectors.
\end{proposition}
\begin{proof}
Let $t$ be a counting node. There are $S_t$ subfunction classes, one per distinct nonzero row of $M_t(f)$. These $S_t$ rows span the row space of $M_t(f)$, which has dimension $\rho_t = S_t$, so they are linearly independent.

We show that the scan keeps $\tau$ if and only if $\tau$ is the lexicographically least member of its class. If $f[\tau] = 0$, then $\tau$ belongs to no class and is discarded. If $f[\tau] \neq 0$ and some $\tau' < \tau$ has $f[\tau'] = f[\tau]$, then $f[\tau']$ is either a kept subfunction or a linear combination of kept subfunctions, and so is $f[\tau]$, which is therefore discarded. Finally, let $\tau$ be the least member of its class and suppose that the claim holds for all assignments before $\tau$. The subfunctions already kept are then those of the least members of other classes, so they are distinct rows, different from $f[\tau]$. By $\rho_t = S_t$, the distinct nonzero rows are linearly independent, so $f[\tau]$ is not a linear combination of them, and $\tau$ is kept. Hence, $p^t_1, \dots, p^t_{\rho_t}$ are the least members of the classes.

Now fix $\tau \in 2^{X_t}$. If $\tau$ belongs to the class of $p^t_l$, then $f[\tau] = f[p^t_l]$, and the uniqueness of the coefficients in (\ref{eq:linear_combination}) gives $e^t_l(\tau) = 1$ and $e^t_i(\tau) = 0$ for $i \neq l$. If $f[\tau] = 0$, the same uniqueness gives $e^t_i(\tau) = 0$ for every $i$. So $e^t_i(\tau) = 1$ if $\tau$ belongs to the class of $p^t_i$ and $e^t_i(\tau) = 0$ otherwise: $e^t_i$ is the~$0/1$ indicator vector of the class of $p^t_i$.
\end{proof}

If $f \not\equiv 0$ and every node is a counting node, the ATDD \emph{is} the canonical minimal
TDD of \mbox{\citet{CapelliEtAl26}}, read as a network via
Proposition~\ref{proposition:simulation}.

\section{Queries and Operations}\label{sec:queries}

We review the tractability of various queries and operations on ATDD. Every algorithm in this section also receives the vtree $T$ as input, since the $0$ ATDD does not determine it. Since the ATDD of a function $f \not\equiv 0$ is a TTN, since the $0$ ATDD translates into the TTN over $T$ with all dimensions $1$ and all entries $0$, and since TTNs computing Boolean functions can be brought to ATDD form in polynomial time, an operation or query on Boolean functions is tractable for ATDD if and only if it is tractable for TTNs computing Boolean functions.

Consider an ATDD representing the Boolean function $f$. The simplest query, satisfiability testing (is $f^{-1}(1)$ non-empty?), takes constant time on ATDD as the only ATDD for zero function is the $0$ ATDD. Determining the existence of falsifying assignments (is $f^{-1}(0)$ non-empty?) is also tractable since model counting, i.e., determining $|f^{-1}(1)|$ is tractable on ATDD. The more general task of weighted model counting, i.e., computing $\sum_{\tau \in f^{-1}(1)}w(\tau)$ given a weight function~$w$ is also tractable for multiplicative weight functions: each literal~$\ell$ over $X$ is associated with a weight $w(\ell)$ and the weight $w(\tau)$ of an assignment $\tau$ is the product of the weights of its literals. Formally, $w(\tau) = \prod_{x \in \tau^{-1}(1)} w(x)\prod_{x \in \tau^{-1}(0)}w(\neg x)$. On ATDD, weighted model counting is done by multiplying each leaf tensor with $(w(\neg x)\ w(x))^{\top}$ and doing the full contraction of the TTN.

\begin{restatable}{theorem}{counting}\label{thm:modelCounting}
Consider the ATDD respecting $T$ for the non-zero Boolean function~$f$ and a multiplicative weight function $w$. There is an algorithm that, given a node $t$ of $T$ and $i \in [d(t)]$, computes $\sum_{\tau \in 2^{X_t}} w(\tau)u^t_i(\tau)$ with a number of arithmetic operations polynomial in the size of the ATDD. For~$t$ the root of $T$ and $i = 1$, it returns $\sum_{\tau \in f^{-1}(1)}w(\tau)$.
\end{restatable}
\begin{proof}
This is clear for $t$ a leaf of $T$ since then $|X_t|=1$. Now say $t$ is an internal node with children~$t_1$ and~$t_2$. For $\tau$ an assignment to $X_t$, $\tau_1$ and $\tau_2$ denote the restrictions to $X_{t_1}$ and~$X_{t_2}$, respectively.
\begin{align*}
\sum_{\tau \in 2^{X_t}} w(\tau)u^t_i(\tau) &= \sum_{\tau \in 2^{X_t}} \sum_{j \in [d(t_1)]}\sum_{k \in [d(t_2)]}W^t_{ijk} w(\tau_1)w(\tau_2)u^{t_1}_j(\tau_1)u^{t_2}_k(\tau_2) 
\\
&=  \sum_{j \in [d(t_1)]}\sum_{k \in [d(t_2)]}W^t_{ijk} \left(\sum_{\tau_1 \in 2^{X_{t_1}}}w(\tau_1)u^{t_1}_j(\tau_1)\right)\left(\sum_{\tau_2 \in 2^{X_{t_2}}}w(\tau_2)u^{t_2}_k(\tau_2)\right)
\end{align*}
With the above equality, the sums $\sum_{\tau} w(\tau)u^t_i(\tau)$ for all $i \in [d(t)]$ are computed once per node, bottom-up, with a number of arithmetic operations polynomial in the size of the ATDD. When $t$ is the root of $T$, $d(t) = 1$ and $f(\tau) = u^t_1(\tau)$, so $\sum_{\tau}w(\tau) u^t_1(\tau)$ is the weighted model count of $f$.
\end{proof}

For ATDDs respecting the same vtree, equivalence testing takes linear time as it boils down to syntactical equality testing (Theorem~\ref{thm:canonical}). 

Conditioning on partial assignments is tractable, and therefore so is determining implicates and implicants, or enumerating models, see the queries called CE, IM and ME in~\citep{Darwiche2002}.

For $Y \subsetneq X$, the vtree $T|Y$ \emph{induced} on $X \setminus Y$ is obtained from $T$ by deleting the leaves labeled by $Y$ together with every node all of whose leaves are deleted, and then contracting every node that is left with a single child (a root left with a single child is replaced by that child); for a single variable we write $T|x$.

\begin{restatable}{theorem}{conditioning}\label{thm:conditioning}
There is a polynomial-time algorithm that, given an ATDD respecting $T$ for the Boolean function~$f$ over $X$ and a partial assignment $\tau$ whose set $\var(\tau)$ of assigned variables is a proper subset of $X$, returns an ATDD representing $f[\tau]$ that respects $T|\var(\tau)$.
\end{restatable}
\begin{proof}
If the input is the $0$ ATDD, then $f[\tau] \equiv 0$ and the $0$ ATDD is returned. Otherwise, consider first a single variable $x$ with value $b$, and let $\ell$ be the leaf of $x$, $p$ its parent and $s$ its sibling; say $\ell$ is the left child of $p$ (otherwise contract along the third index below). Multiplying the leaf matrix~$u^\ell$ by the unit vector of $b$ gives the column vector $v \in \Q^{d(\ell)}$ with $v_j = u^\ell_j(b)$, and by \eqref{eq:evaluate_ttn} the bond vectors at $p$ become $u^p_i = \sum_k G_{ik}\, u^s_k$ with $G_{ik} = \sum_j W^p_{ijk}\, v_j$. So $p$ collapses onto $s$. We delete~$\ell$ and $p$, replace $\bar W^s$ by $G\, \bar W^s$ (at a leaf, $u^s$ by $G\, u^s$) so that~$s$ now has $d(p)$ bond vectors, and let the parent of $p$ point at $s$ (if $p$ is the root, $s$ becomes the root, with $d(p) = 1$). The result is a TTN respecting $T|x$ that computes $f[x = b]$, with every dimension at most the width $w$ of the input. It is in general neither minimal nor in echelon form, since the cut matrix of $f[x=b]$ at a node is a submatrix of the cut matrix of~$f$, so no rank increases, but ranks can drop at every node, for instance when $f = x \vee (y \wedge z)$ over the vtree $(x,(y,z))$ and $b = 1$, where the result is the constant $1$ and every cut rank drops to $1$. Proposition~\ref{prop:minimization} then yields a minimal TTN, and Theorem~\ref{thm:makeCanonical} the ATDD of $f[x = b]$, both in polynomial time (if $f[x = b] \equiv 0$, the minimization detects it and we return the ATDD of the~$0$ function; when the result is a single leaf, its $1 \times 2$ leaf matrix is inspected directly; it is minimal unless zero). For a partial assignment $\tau$, repeat this for each variable of $\tau$ in turn, stopping with the $0$ ATDD as soon as an intermediate result is zero; every intermediate ATDD has width at most $w$, and the entries of an ATDD of width $w$ have bit size $O(w \log w)$ (\suppl), so the total cost is polynomial.
\end{proof}

For classical Boolean operations on ATDDs, we can express each operation in terms of multilinear algebra and use known algorithms for manipulating TTNs. Given two functions $f, g: 2^X \rightarrow \mathbb{Q}$ and two TTNs for $f$ and $g$ that both respect the same vtree $T$, there is an algorithm computing a TTN for the sum $f + g$ and another algorithm computing a TTN for the entrywise product (or Hadamard product) $f \odot g$, both running in polynomial time. The Hadamard product takes at every leaf the entrywise products of the pairs of bond vectors and at every internal node the Kronecker product of the two transfer tensors, so the bond dimensions multiply. The sum stacks the two leaf matrices at every leaf and places the two transfer tensors block-diagonally at every internal node, so the bond dimensions add, except at the root, where $d(r) = 1$ and the two root tensors become the two diagonal blocks of one root tensor with a single row; if the root is a leaf, the two leaf vectors are added instead. Scaling turns the function $f$ into $\lambda f$ for some $\lambda \in \mathbb{Q}$; a TTN can be scaled in polynomial time.

If $f$ and $g$ are Boolean functions, $f, g: 2^X \rightarrow \{0,1\}$, then conjunction, disjunction and negation become $f \odot g = f \land g$, $f \lor g = f + g - f \odot g$ and $\neg f = 1 - f$. Given the ATDDs $A_f$ and $A_g$ for $f$ and $g$ that both respect $T$, computing TTNs for these expressions is done in polynomial time; the resulting TTNs are brought into ATDD form using Proposition~\ref{prop:minimization} and Theorem~\ref{thm:makeCanonical}. Single-variable forgetting, i.e., computing $\exists x.f = f[x=0] \lor f[x = 1]$, is also tractable by Theorems~\ref{thm:conditioning} and~\ref{thm:operations}.   

\begin{theorem}\label{thm:operations}
There are polynomial-time algorithms that, given two ATDDs for the Boolean functions~$f$ and $g$ that both respect the vtree $T$, return ATDDs representing $f \land g$, $f \lor g$ and $\neg f$ that all respect $T$.
\end{theorem}
\begin{proof}
If $A_f$ is the $0$ ATDD, the algorithms return the $0$ ATDD for $f \wedge g$, the input $A_g$ for $f \vee g$, and for $\neg f$ the ATDD of the constant $1$, which has $d(t) = 1$ at every node of $T$, leaf matrices $(1\ 1)$ and transfer tensors $(1)$; if $A_g$ is the $0$ ATDD, symmetrically for $f \wedge g$ and $f \vee g$. Otherwise, on Boolean inputs $f \wedge g = f \odot g$ is the Hadamard product of the two TTNs, with dimensions $d_f(t)\, d_g(t)$; $\neg f = 1 - f$ is the sum of the all-ones TTN (all dimensions $1$) and $A_f$ scaled by $-1$, with dimensions $d_f(t) + 1$ below the root; and $f \vee g = f + g - f \odot g$ is two sums and one Hadamard product. Each is a TTN respecting $T$ that computes a Boolean function, of width polynomial in the input widths (the rank bounds in Table~\ref{tab:ops}), and Proposition~\ref{prop:minimization} followed by Theorem~\ref{thm:makeCanonical} turns it into the ATDD in polynomial time; if the function is zero, which the minimization detects, the $0$ ATDD is returned.
\end{proof}

\begin{theorem}\label{thm:forgetting}
There is a polynomial-time algorithm that, given an ATDD for the Boolean function~$f$ over $X$ that respects the vtree $T$ and a variable $x \in X$ with $|X| \ge 2$, returns an ATDD representing~$\exists x.f$ that respects $T|x$.
\end{theorem}
\begin{proof}
$\exists x. f = f[x = 0] \vee f[x = 1]$: two applications of Theorem~\ref{thm:conditioning}, both yielding ATDDs respecting $T|x$, followed by the disjunction of Theorem~\ref{thm:operations}.
\end{proof}

The algebra on TTN also allows us to check whether a given TTN represents a Boolean function.
\begin{restatable}[$\dagger$]{theorem}{promise}\label{thm:promise}
A TTN computing $f$ satisfies the Boolean promise if and only if
\begin{equation}\label{eq:booleanCriterion}
\sum_{\tau \in 2^X} (f(\tau)^2 - f(\tau))^2 = 0.
\end{equation}
\end{restatable}
\begin{proof}
A sum of squares of rationals is $0$ if and only if every term is $0$, and $f(\tau)^2 = f(\tau)$ holds if and only if $f(\tau) \in \{0,1\}$.
\end{proof}

The Boolean promise is therefore decidable in polynomial time. Let $w$ be the width of the given TTN. With the sum and Hadamard product of TTNs, whose dimensions are given in the proof of Theorem~\ref{thm:operations}, we obtain a TTN for $(f^2 - f)^2 = f \odot f \odot f \odot f - 2\, f \odot f \odot f + f \odot f$ of width at most $w^4 + w^3 + w^2$ in polynomial time. The contraction in the proof of Theorem~\ref{thm:modelCounting} applies verbatim to any TTN. With the weight $1$ on every literal it computes $\sum_{\tau} (f(\tau)^2 - f(\tau))^2$ with a number of arithmetic operations polynomial in the size of that TTN. By Theorem~\ref{thm:promise},~$f$ is Boolean if and only if this value is $0$.

\begin{table}[t]
\centering
\small
\setlength{\tabcolsep}{4pt}
\begin{tabular}{@{}lll@{}}
\toprule
Operation & Result & Rank bound at every $t$\\
\midrule
conjunction & $f \wedge g$ & $\rho_t(f)\,\rho_t(g)$\\
negation & $\neg f$ & $\rho_t(f) + 1$\\
disjunction & $f \vee g$ & $\rho_t(f)\rho_t(g) {+} \rho_t(f) {+} \rho_t(g)$\\
conditioning & $f[\tau]$ & $\rho_t(f)$\\
forgetting ($1$ var) & $\exists x\, f$ & $\rho_t(f)^2 + 2\rho_t(f)$\\
\bottomrule
\end{tabular}
\vspace{-6pt}
\caption{Various operations and tests and their rank bound at each
node $t$. Conditioning is read at the corresponding node of the induced vtree.}
\label{tab:ops}
\end{table}

\section{Vtree Rotation and Vtree Change}\label{sec:rotation}

All operations so far fix the vtree. \citet{CapelliEtAl26} ask for the complexity of passing from one vtree to another, measured in the input and the output, and suggest adapting the local vtree changes of SDD compilers~\citep{ChoiD13} to find a good vtree. We answer the question for TDDs and for ATDDs, and we show that a rotation changes only two tensors of the ATDD, which are computed from the two old ones.
\citet{ZhangWangArenasBroeck25} study vtree change for probabilistic circuits.

A left rotation rewrites a node $v = (a,t = (b,c))$ as $v' = (s = (a,b),c)$. The cut ranks for $a$, $b$, $c$ and $v$ are unchanged, but the cut for $t$ is replaced by the cut for $s$. The cut rank $\rho_s$ for $s$ is at most $\min(\rho_a\rho_b,\ \rho_c\rho_v)$, with $\rho_a, \rho_b, \rho_c$ the cut ranks for $a, b, c$ and $\rho_v$ the cut rank for $v$; the two terms come from the two sides of the cut, and the bound is tight (\supplb). A single rotation thus increases the factor rank to at most its square.

\begin{figure}[h]
\centering
\begin{tikzpicture}[scale=1.0]
\node[circle,inner sep=2pt,draw,label={west:\scriptsize $v$}] (v) at (0,0) {};
\node[circle,inner sep=2pt,draw,label={west:\scriptsize $a$}] (a) at (-0.5,-1) {};
\node[circle,inner sep=2pt,draw,label={east:\scriptsize $t$}] (t) at (+0.5,-1) {};
\node[circle,inner sep=2pt,draw,label={west:\scriptsize $b$}] (b) at (+0,-2) {};
\node[circle,inner sep=2pt,draw,label={east:\scriptsize $c$}] (c) at (+1,-2) {};

\draw (a) -- (v) -- (t);
\draw (b) -- (t) -- (c);

\def\o{5}

\node[circle,inner sep=2pt,draw,label={west:\scriptsize $v'$}] (v) at (\o+0,0) {};
\node[circle,inner sep=2pt,draw,label={west:\scriptsize $a$}] (a) at (\o-1,-2) {};
\node[circle,inner sep=2pt,draw,label={west:\scriptsize $s$}] (s) at (\o-0.5,-1) {};
\node[circle,inner sep=2pt,draw,label={east:\scriptsize $b$}] (b) at (\o+0,-2) {};
\node[circle,inner sep=2pt,draw,label={east:\scriptsize $c$}] (c) at (\o+0.5,-1) {};

\draw (s) -- (v) -- (c);
\draw (a) -- (s) -- (b);

\draw[-latex] (0.75,-0.75) to[out=45,in=135] node[above,midway]{left rotation} (\o-0.75,-0.75);
\end{tikzpicture}
\end{figure}

The rotation modifies only the tensor for $v$ and that for $s$. The canonical tests for $v$ are unchanged since they depend only on the variables below $v$ and the variable ordering induced by the vtree, both of which are preserved. The canonical tests for $s$ are products of the canonical tests for $a$ and $b$.

The rotation proceeds as follows: first we contract $W^v$ and $W^t$ over the index of $t$ and get the $d(v) \times d(a) \times d(b) \times d(c)$ tensor $\Theta$ with $\Theta_{ijkl} = \sum_{m \in [d(t)]} W^v_{ijm} W^t_{mkl}$, next we get the $d(v)d(c) \times d(a)d(b)$ matrix $R$ defined by $R_{(i,l),(j,k)} = \Theta_{i,j,k,l}$ and compute its reduced row echelon form $R'$ with pivot columns indexed $(j_1,k_1), \dots, (j_\rho,k_\rho)$, finally we set $d(s) = \rho$ and create the $d(s) \times d(a) \times d(b)$ tensor for $s$ and the $d(v) \times d(s) \times d(c)$ tensor for $v'$ as
$$
W^s_{h,j,k} = R'_{h,(j,k)} \quad \text{ and } \quad W^{v'}_{i,h,l} =  \Theta_{i,j_h,k_h,l}.
$$
The contraction costs $d(v)d(a)d(b)d(c)d(t)$ multiplications; computing the reduced row echelon form of $R$ takes $O(\text{rows}\cdot \text{cols} \cdot \text{rank}) = O(d(v)d(a)d(b)d(c)d(s))$ arithmetic operations. So the rotation costs $O(d(v)d(a)d(b)d(c)(d(t)+d(s)))$.

\begin{theorem}[$\star\dagger$]\label{thm:rotation}
Let $f \not\equiv 0$ be a Boolean function and let $T'$ be obtained from $T$ by a left rotation at $v$. The ATDD of $f$ for $T'$ differs from the ATDD of $f$ for $T$ only in the two affected tensors. At the new node $s$, the canonical tests are products of the canonical tests at~$a$ and $b$, and $\rho_s$ is the rank of the matrix $R$ above.
\end{theorem}

The procedure above returns the ATDD of $f$ for $T'$ (\supplb).

Any vtree $T_2$ is reachable from a vtree $T_1$ in $O(n^2)$ rotations and swaps (exchanging the two children of a node). A rotation is local. A swap is free on the tensors (exchange the two child index groups) and keeps every cut rank, so the swapped TTN is still minimal and Theorem~\ref{thm:makeCanonical} re-canonicalizes it in polynomial time, but the canonical tests may change at the swapped node and all its ancestors, so a swap is not local. We do not know whether such a path can avoid wide intermediate ATDDs (Section~\ref{sec:conclusion}), so in the next theorem we avoid the path and build the target directly. We assume we have a TTN $A_1$ respecting $T_1$ and computing $f$.

The construction processes the nodes of $T_2$ bottom-up, identifying a node with its variable set~$Y$, and builds for each $Y$ its canonical tests and its tensor. By Lemma~\ref{lem:compose}, applied to~$T_2$, every canonical test at an internal $Y$ is a product $p^{Y_1}_j \times p^{Y_2}_k$ of tests of the children, so there are at most $\rho(Y_1)\rho(Y_2)$ candidates (at a leaf, the two values of its variable), and the greedy scan of Section~\ref{sec:canon} over these candidates alone returns exactly the tests at $Y$. 
During the scan, a candidate $\tau$ is kept iff $f[\tau]$ is not in the span of the rows $\tau_1,\dots,\tau_k$ kept so far. This is decided with the Gram matrix~$G$ of the kept rows, i.e., the matrix whose $(\tau_i,\tau_j)$ entry is $\langle f[\tau_i], f[\tau_j] \rangle$. The $f[\tau_1],\dots,f[\tau_k]$ are linearly independent iff $\det(G) \neq 0$ so constructing the Gram matrix during the scan allows us to determine whether to keep $\tau$ as a canonical test. Once the tests $p^Y_1, \dots, p^Y_\rho$ at $Y$ are known, the tensor entries $W^Y_{ijk} = e^Y_i(p^{Y_1}_j \times p^{Y_2}_k)$ are the coefficients of each candidate row in the test rows, which solve $G\,e(\tau) = b(\tau)$ with $b_l(\tau) = \langle f[\tau], f[p^Y_l] \rangle$, so the Gram matrix also allows us to construct our tensors. Each Gram value $\langle f[\tau], f[\tau'] \rangle$ is a contraction of two copies of $A_1$ in which the variables of $Y$ are fixed to the values prescribed by $\tau$ and $\tau'$. So we access $A_1$ only through inner products of subfunctions and write out no subfunction. The cost is $O(n^2 w_1^4 w_2^4)$ arithmetic operations, with $w_1$ the width of~$A_1$ and $w_2 = \fr(f,T_2)$ discovered bottom-up. Each Gram value costs $O(n w_1^4)$ arithmetic operations, and a node of $T_2$ needs $O(w_2^4)$ of them. The determinant test is done incrementally, by extending a factorization of $G$ by one row per kept test, in $O(k^2)$ operations per candidate. All numbers have bit size polynomial in $n$, $w_1$, $w_2$ and the bit size of the entries of $A_1$ (\supplb).

\begin{theorem}[$\star\dagger$]\label{thm:change}
Let $A_1$ be a network respecting $T_1$ and computing a Boolean function $f
\not\equiv 0$, and let $T_2$ be any target vtree over $X$. A single bottom-up
pass over $T_2$ yields the ATDD of $f$ for $T_2$, selecting at each node the canonical tests among the products of the children's tests.
\end{theorem}

The pass is output-polynomial: it discovers the target width $w_2 = \fr(f, T_2)$
bottom-up, with no a priori bound on $w_2$, and stops as soon as the width
exceeds a given bound. The same
argument, with subfunction classes in place of tests, produces the canonical
minimal TDD for
$T_2$ in time polynomial in the size of the input TDD, $n$, and the output width (\supplb); for a single rotation, with the canonical minimal TDD for $T_1$ as input, it can be restricted to
the two affected nodes, an exact local rotation for TDDs. This answers the
complexity question of \citet{CapelliEtAl26}. Like the ATDD rotation, the TDD rotation is polynomial in the input, since $S_s \le S_a S_b$; but counting has no analogue of the complement-side bound $\rho_s \le \rho_c\rho_v$, as the multiplexer already shows (\supplb).

\section{Validation}\label{sec:validation}

\paragraph{Twin compilers.}
We implemented two prototype compilers in Rust, one for minimal TDDs and one for
ATDDs. Both compilers share the same infrastructure layer (DIMACS
parsing, vtree construction with a min-fill heuristic, assignment indexing);
consequently, both run over identical vtrees. The compile loop follows
Algorithm~1 of \citet{CapelliEtAl26}: we start from the constant-$1$ diagram,
conjoin the clause diagrams one at a time, and reduce after each. The compilers
differ in the conjoin and reduce steps. The TDD compiler classifies subfunctions and
rebuilds the diagram. The ATDD compiler applies a Hadamard product, minimizes each bond
dimension to its rank, and canonicalizes to the echelon basis.
All ATDD arithmetic is exact over $\Q$. Compilation and model counting use the
\emph{integer kernel} described below. The ATDD algorithms (bond-dimension
minimization, canonicalization, operations, promise test, rotation surgery)
are checked by exact-arithmetic differential testing against brute-force
references and an independent Python oracle. Both compilers, the benchmark
scripts, and the data behind every number we report are in the Zenodo repository.

\paragraph{Integer kernel.}
All the numbers in the diagram are fractions. At each node the denominators all
divide one number. Hence, we multiply each node tensor by this number, which
gives us the entire diagram with integers instead of rationals. The expensive
inner loops, which dominate compiling and counting time, can now run on this
integer variant. They return values that are multiplied by a positive factor. It
is the same factor for every entry. But checking which columns are pivots and
whether an entry is zero does not depend on such a factor. We still store and transform
the rational diagram, and we divide each value taken from the integer variant by its known
factor or use it only in a ratio in which the factor cancels. Consequently, the
canonical diagram constructed is exactly the same, entry by entry. The implementation
keeps the product of the numbers by which it multiplied the node tensors and
divides every model count by it, so the counts are exact.

\paragraph{Empirical results.}
The timings are preliminary. All timed experiments ran on an Apple M5 (4 performance and 6 efficiency cores, 32\,GB RAM, macOS~26.6; rustc~1.99.0);
core-benchmark compile times are medians of up to three runs, counting
times are means over repeated queries, and the time caps were $300$
seconds for the core benchmarks and $30$ minutes for the single-run
scaling sweep.
We ran tests across $1{,}632$ cuts over six families of functions, with
$4$ or $5$ vtrees per instance. Each single cut satisfied $\rho_t \le S_t$, and
every entry of the computed ATDDs required not more than three bits. The gap between count
and rank appeared mainly on the gap families $\MUX$ and $\HWB$: for the multiplexer
with eight data variables, the largest count was $255$ against a largest rank of~$8$; on the hidden weighted bit
function at $n = 10$, count $31$ and rank $10$. Among the other families, only the random $3$-CNFs showed a gap, at $27$ of their $304$ cuts. On eight native CNFs, the model counts of both compilers
agreed with the independent exact counter sharpSAT-TD \citep{Korhonen2021}.

\begin{figure}[t]
\centering
\includegraphics[width=\wscaling\columnwidth]{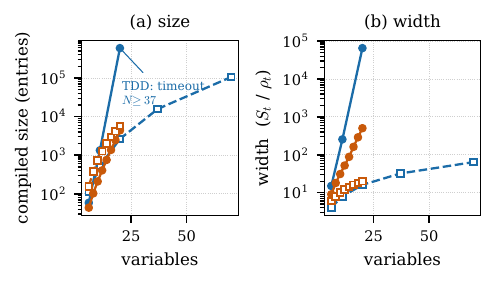}
\vspace{-17pt}
\caption{Empirical behavior on the gap families
(\textcolor{figmux}{$\blacksquare$}~$\MUX$,
\textcolor{fighwb}{$\blacksquare$}~$\HWB$; solid lines are TDDs, dashed
lines are ATDDs): \emph{(a)} compiled size and \emph{(b)} width ($S_t$ for
the TDD, $\rho_t$ for the ATDD) against the number of variables ($N$ for
$\MUX$, $n$ for $\HWB$), log scale. The multiplexer TDD curve ends at~$N =
20$; the next instance, $N = 37$, exceeded the $30$-minute compile timeout.}
\label{fig:scaling}
\end{figure}

\paragraph{Asymptotic separation.}
The advantage of ATDDs (Theorem~\ref{thm:separation}) shows in practice when
the instance size increases and where the two width measures diverge
(Figure~\ref{fig:scaling}). On the
multiplexer, over vtrees with a node whose variables are exactly the data variables, as in our runs, the minimal TDD width is the subfunction count $S_t = 2^{2^k} - 1$,
exponential in the number $2^k$ of data variables, whereas the ATDD width is the
rank $\rho_t = n = 2^k$, with $N = n + k$ variables in total. At $N = 20$, that
is, $16$ data and $4$ selector variables, the TDD already has width
$65{,}535$, compared to an ATDD of width $16$. On a suitable vtree the multiplexer has a TDD of linear size \citep{CapelliEtAl26}. At $N = 37$, the TDD cannot be compiled within $30$
minutes; the ATDD compiles in $5$ seconds and counts all $2^{36}$ models in~$14\,\mu$s. At $N = 70$, the TDD times out again, while the ATDD compiles in $5$ minutes
and counts all~$2^{69}$ models in $72\,\mu$s. We see a similar picture for the hidden weighted bit function: the TDD
width increases to $503$ at $n = 20$, while the ATDD width stays at $n$ on the vtrees of our runs.

\paragraph{ATDD overhead cost.}
Since each ATDD node stores a rational
tensor, the compiled ATDD is larger than the corresponding TDD of the same
width. On the core benchmark tier, the rational model counting is $24$ to
$49$ times slower than the TDD's machine-integer counting (geometric means per family); the integer kernel
removes most of this overhead, bringing counting to within a factor of $1.3$ to $2.1$. ATDD
compilation is likewise slower than that of the TDD. On the pigeonhole and random~$3$-CNF
instances we tested, the TDD width and ATDD width coincide at almost every node;
there, ATDDs have no advantage and incur a larger overhead, and on the
larger pigeonhole instances the ATDD compiler exceeded the time cap that
the TDD met.

\section{Conclusion}\label{sec:conclusion}

We introduced ATDDs by replacing the subfunction counting of TDDs by
cut-matrix rank, keeping the properties of TDDs (canonicity, the operation set, and a width equal to a parameter of the function and the vtree) while gaining super-polynomial succinctness.
Two problems remain open: whether the quasi-polynomial d-SDNNF separation
can be sharpened to an exponential one, and whether every two low-rank vtrees are joined by a
rotation path without high-rank intermediates, since a
negative answer means that rotation-based vtree search can get stuck.

\subsection*{Use of AI Assistance}

Parts of the paper, the supplement, the Lean formalization and the code were written with substantial assistance from large language models, under the authors' direction. The statements of the paper marked with a dagger are proved in Lean, whose kernel checks a proof regardless of how it was produced. The algorithms and explicit bounds are tested by exact-arithmetic scripts in the Zenodo repository. The authors checked the natural-language proofs, which were also audited by further language-model runs. The authors take full responsibility for all content.

\subsection*{Acknowledgments}

This research was funded in part by the Austrian Science Fund (FWF) within the Cluster of Excellence Bilateral Artificial Intelligence (10.55776/COE12) and the project STRIDES (10.55776/P36420).

\bibliography{refs}

\providecommand{\stepl}[1]{\smallskip\noindent\textbf{#1}}

\appendix

\section{Proof of Proposition~\ref{proposition:nestedness}}\label{app:nested}

Throughout, $P$ and $Q$ are finite index sets. A vector $a \in \Q^P$ is a
column of $|P|$ numbers indexed by $P$, and a matrix $N \in \Q^{P \times Q}$
has rows indexed by $P$ and columns indexed by $Q$. We write $N(\eta, \cdot)
\in \Q^{Q}$ for the row of $N$ at $\eta \in P$ and $N(\cdot, \kappa) \in
\Q^{P}$ for the column at $\kappa \in Q$.

\stepl{Vectors over $X_t$ are matrices.}
Let $t$ be an internal vtree node with children $t_1$ and $t_2$, so $X_t$ is
the disjoint union of $X_{t_1}$ and $X_{t_2}$. An assignment $\tau \in
2^{X_t}$ is the same thing as a pair $(\eta, \kappa)$ with $\eta \in
2^{X_{t_1}}$ and $\kappa \in 2^{X_{t_2}}$, namely $\tau = \eta \times \kappa$.
Hence a vector $v \in \Q^{2^{X_t}}$ is the same thing as a matrix with rows
indexed by $2^{X_{t_1}}$ and columns indexed by $2^{X_{t_2}}$, the entry at
$(\eta, \kappa)$ being $v(\eta \times \kappa)$. We use both views of $v$
freely and do not distinguish them in notation.

\stepl{Products are matrices of rank at most one.}
For $a \in \Q^{P}$ and $b \in \Q^{Q}$, the product $a \otimes b \in \Q^{P
\times Q}$ has entries $(a \otimes b)(\eta, \kappa) = a(\eta)\, b(\kappa)$;
for $P = 2^{X_{t_1}}$ and $Q = 2^{X_{t_2}}$ this is exactly the product of
Section~\ref{sec:prelim}. In matrix language, $a \otimes b$ is the outer
product $a\, b^{\top}$: its column at $\kappa$ is $b(\kappa) \cdot a$, a
multiple of $a$, and its row at $\eta$ is $a(\eta) \cdot b$, a multiple of
$b$. So $a \otimes b$ has rank at most one, and conversely every matrix of
rank at most one is of this form.

\stepl{The tensor product of two subspaces.}
For subspaces $U \subseteq \Q^{P}$ and $W \subseteq \Q^{Q}$ we define
\[
  U \otimes W \;=\; \operatorname{span}\{\, a \otimes b : a \in U,\ b \in W \,\}
  \;\subseteq\; \Q^{P \times Q}.
\]
Since a scalar factor can be absorbed into $a$, this is the set of all
finite sums $\sum_i a_i \otimes b_i$ with $a_i \in U$ and $b_i \in W$. In
matrix language, $U \otimes W$ consists of the matrices that can be written
as a sum of rank-one matrices whose column factor lies in $U$ and whose row
factor lies in $W$. With $U = V_{t_1}$ and~$W = V_{t_2}$ this is the space
$V_{t_1} \otimes V_{t_2}$ of Proposition~\ref{proposition:nestedness}.

\stepl{A necessary condition.}
Let $N = \sum_i a_i \otimes b_i$ with $a_i \in U$ and $b_i \in W$. Reading
off one column and one row of this sum,
\[
  N(\cdot, \kappa) = \sum_i b_i(\kappa)\, a_i \in U
  \qquad\text{and}\qquad
  N(\eta, \cdot) = \sum_i a_i(\eta)\, b_i \in W .
\]
So every matrix in $U \otimes W$ has all its columns in $U$ and all its rows
in $W$. The matrix lemma below states that the converse holds as well. That
converse is the only piece of algebra in the proof of nestedness.

\begin{lemma}[$\dagger$]\label{lem:app-sandwich}
Let $U \subseteq \Q^{P}$ and $W \subseteq \Q^{Q}$ be subspaces and let $N
\in \Q^{P \times Q}$ be a matrix all of whose columns lie in $U$ and all of
whose rows lie in $W$. Then $N \in U \otimes W$.
\end{lemma}

\begin{proof}
Let $m = \rk N$. If $m = 0$ then $N = 0 \in U \otimes W$, so assume $m \ge 1$.

\stepl{Step 1: an invertible $m \times m$ submatrix.}
Choose $m$ linearly independent columns of $N$, at indices $\kappa_1, \dots,
\kappa_m \in Q$, and let $C \in \Q^{P \times m}$ be the matrix formed by
these columns; the columns of $C$ are a basis of the column space of $N$.
Since $C$ has rank $m$ and row rank equals column rank, $C$ also has~$m$
linearly independent rows, at indices $\eta_1, \dots, \eta_m \in P$. Let $S
\in \Q^{m \times m}$ be the submatrix of $N$ on these rows and columns,
$S_{ji} = N(\eta_j, \kappa_i)$, and let $R \in \Q^{m \times Q}$ be the
matrix formed by the rows $\eta_1, \dots, \eta_m$ of $N$. The rows of $S$
are the selected rows of $C$, so $S$ has rank $m$ and is invertible.

\stepl{Step 2: a column is determined by its entries in the selected rows.}
Every column of $N$ is a linear combination of the columns of $C$, so $N =
C\, Y$ for the matrix $Y \in \Q^{m \times Q}$ whose column at $\kappa$
holds the coefficients that express $N(\cdot, \kappa)$ in terms of the
selected columns. Restricting the identity~$N = C\,Y$ to the rows $\eta_1,
\dots, \eta_m$ gives $R = S\, Y$. Hence $Y = S^{-1} R$ and
\begin{equation}\label{eq:skeleton}
  N \;=\; C\, S^{-1} R .
\end{equation}

\stepl{Step 3: read \eqref{eq:skeleton} as a sum of products.}
Writing $c_i = N(\cdot, \kappa_i)$ for the selected columns and $r_j =
N(\eta_j, \cdot)$ for the selected rows, the product $C\, S^{-1} R$ expands
entrywise to
\[
  N \;=\; \sum_{i=1}^{m} \sum_{j=1}^{m} (S^{-1})_{ij}\; c_i \otimes r_j .
\]
By hypothesis $c_i \in U$ and $r_j \in W$. So $N$ is a linear combination of
products $a \otimes b$ with $a \in U$ and~$b \in W$, that is, $N \in U
\otimes W$.
\end{proof}

\lemnested*

\begin{proof}
Recall from Section~\ref{sec:prelim} that
$V_t = \operatorname{span}\{ f[\sigma] : \sigma \in 2^{X \setminus X_t} \}$.
The right-hand side $V_{t_1} \otimes V_{t_2}$ is a subspace, closed under
sums and scalar multiples, so it suffices to show $f[\sigma] \in V_{t_1}
\otimes V_{t_2}$ for each generator $f[\sigma]$.

Fix $\sigma \in 2^{X \setminus X_t}$ and view $f[\sigma] \in \Q^{2^{X_t}}$
as the matrix $N_\sigma$ with rows indexed by $2^{X_{t_1}}$, columns indexed
by $2^{X_{t_2}}$, and entries
\[
  N_\sigma(\eta, \kappa) \;=\; f[\sigma](\eta \times \kappa)
  \;=\; f(\eta \times \kappa \times \sigma).
\]

\stepl{The columns of $N_\sigma$ lie in $V_{t_1}$.}
Fix $\kappa \in 2^{X_{t_2}}$. The column $N_\sigma(\cdot, \kappa)$ is the
vector $\eta \mapsto f(\eta \times \kappa \times \sigma)$ on $2^{X_{t_1}}$.
Now $\kappa \times \sigma$ assigns exactly the variables in $X_{t_2} \cup (X
\setminus X_t) = X \setminus X_{t_1}$, so this column is the subfunction
$f[\kappa \times \sigma]$ with $\kappa \times \sigma \in 2^{X \setminus
X_{t_1}}$. Subfunctions of this form are generators of $V_{t_1}$. Hence $N_\sigma(\cdot, \kappa) \in V_{t_1}$.

\stepl{The rows of $N_\sigma$ lie in $V_{t_2}$.}
Symmetrically, fix $\eta \in 2^{X_{t_1}}$. The row $N_\sigma(\eta, \cdot)$
is the vector $\kappa \mapsto f(\eta \times \kappa \times \sigma)$ on
$2^{X_{t_2}}$, which is the subfunction $f[\eta \times \sigma]$ with $\eta
\times \sigma \in 2^{X \setminus X_{t_2}}$, a generator of $V_{t_2}$. Hence
$N_\sigma(\eta, \cdot) \in V_{t_2}$.

\stepl{Conclusion.}
Lemma~\ref{lem:app-sandwich}, applied with $P = 2^{X_{t_1}}$, $Q =
2^{X_{t_2}}$, $U = V_{t_1}$ and $W = V_{t_2}$, gives $N_\sigma \in V_{t_1}
\otimes V_{t_2}$. Under the identification of vectors over $X_t$ with
matrices, $N_\sigma$ is $f[\sigma]$, and the proposition follows.
\end{proof}

\begin{lemma}[$\dagger$]\label{lem:app-basis}
If $a_1, \dots, a_m$ is a basis of $U \subseteq \Q^{P}$ and $b_1, \dots,
b_n$ is a basis of $W \subseteq \Q^{Q}$, then the~$mn$ products $a_j
\otimes b_k$ form a basis of $U \otimes W$.
\end{lemma}

\begin{proof}
\stepl{Spanning.} Every generator $a \otimes b$ of $U \otimes W$ expands as
$a = \sum_j \lambda_j a_j$ and $b = \sum_k \mu_k b_k$, and
$(a \otimes b)(\eta, \kappa) = a(\eta)\, b(\kappa) = \sum_{j,k} \lambda_j
\mu_k\, a_j(\eta)\, b_k(\kappa)$, so $a \otimes b = \sum_{j,k} \lambda_j
\mu_k\, a_j \otimes b_k$.

\stepl{Independence.} Suppose $\sum_{j,k} \alpha_{jk}\, a_j \otimes b_k =
0$. Let $A \in \Q^{P \times m}$ have columns $a_1, \dots, a_m$ and $B \in
\Q^{Q \times n}$ have columns $b_1, \dots, b_n$; then the sum is the matrix
product $A\, \alpha\, B^{\top}$ with $\alpha = (\alpha_{jk}) \in \Q^{m
\times n}$. Since the columns of $A$ are linearly independent, $A$ has a
left inverse $A^{+}$ with $A^{+} A = I_m$: the $m \times m$ matrix $A^{\top}
A$ is invertible, since $z^{\top} A^{\top} A z = \sum_{\eta} (Az)(\eta)^2 > 0$
for every $z \ne 0$, and $A^{+} = (A^{\top} A)^{-1} A^{\top}$ works. Likewise
$B^{+} B = I_n$. Multiplying $A\, \alpha\,
B^{\top} = 0$ by $A^{+}$ on the left and by $(B^{+})^{\top}$ on the right
gives $\alpha = 0$.
\end{proof}

\end{document}